\documentclass{article}
\usepackage[utf8]{inputenc}
\usepackage{physics,amsmath,amssymb,graphicx,url}
\begin{document}
\title{Reconstruction of Sparse Signals using Likelihood Maximization from Compressive Measurements with Gaussian and Saturation Noise}
\author{Shuvayan Banerjee, Radhendushka Srivastava, Ajit Rajwade}
\date{}
\maketitle
\begin{abstract}
Most compressed sensing algorithms do not account for the effect of saturation in noisy compressed measurements, though saturation is an important consequence of the limited dynamic range of existing sensors. The few algorithms that handle saturation effects either simply discard saturated measurements, or impose additional constraints to ensure consistency of the estimated signal with the saturated measurements (based on a known saturation threshold) given uniform-bounded noise. In this paper, we instead propose a new data fidelity function which is directly based on ensuring a certain form of consistency between the signal and the saturated measurements, and can be expressed as the negative logarithm of a certain carefully designed likelihood function. Our estimator works even in the case of Gaussian noise (which is unbounded) in the measurements. We prove that our data fidelity function is convex. We moreover, show that it satisfies the condition of Restricted Strong Convexity and thereby derive an upper bound on the performance of the estimator. We also show that our technique experimentally yields results superior to the state of the art under a wide variety of experimental settings, for compressive signal recovery from noisy and saturated measurements.  
\end{abstract}
\section{Introduction}
Compressed sensing (CS) aims to recover a signal $\boldsymbol{x} \in \mathbb{R}^n$ from its `compressive measurements' of the form $\boldsymbol{y} = \boldsymbol{A x} + \boldsymbol{\eta}$ where $\boldsymbol{A} \in \mathbb{R}^{m \times n}, m \ll n$, is a sensing matrix representing the forward model of the compressive device, and $\boldsymbol{y} \in \mathbb{R}^m$ is a vector of (possibly noisy) compressive measurements. The noise vector is $\boldsymbol{\eta} \in \mathbb{R}^m$. Although this problem is ill-posed for most vectors in $\mathbb{R}^n$, CS theory states that it is well-posed and that the signal $\boldsymbol{x}$ can be recovered with high accuracy \cite{Candes2008}, if $\boldsymbol{x}$ is a sparse (or weakly-sparse) vector, and $\boldsymbol{A}$ obeys the so-called restricted isometry property (RIP). A sensing matrix $\boldsymbol{A}$ is said to obey the RIP of order $s$, if for any $s$-sparse vector $\boldsymbol{x}$, we have $\|\boldsymbol{A x}\|^2_2 \approx \|\boldsymbol{x}\|^2_2$. Here, the degree of approximation is given by the so-called $s$-order restricted isometry constant (RIC) of $\boldsymbol{A}$. There exist precise error bounds for the recovery of $\boldsymbol{x}$ \cite{Candes2008}. Moreover, most of the algorithms for CS recovery are also efficient in terms of computation speed, a well-known example being the LASSO \cite{THW2015}, which seeks to minimize the objective function $J(\boldsymbol{x}) \triangleq \|\boldsymbol{y}-\boldsymbol{A x}\|^2_2 + \lambda \|\boldsymbol{x}\|_1$, given a regularization parameter $\lambda$.

However, the vast majority of the literature assumes a zero mean i.i.d. Gaussian distribution (with known variance) as the noise model. Many practical sensing systems, on the other hand, innately enforce noise of other distributions. Almost all sensors have a fixed (and usually known) dynamic range $[a,b], a < b$. However the underlying signal may be such that not all measurements $\boldsymbol{A^i x}$ (where $\boldsymbol{A^i}$ is the $i^{\textrm{th}}$ row of $\boldsymbol{A}$) can be accommodated within this range. Such measurements then get `clipped' to the value $a$ if $\boldsymbol{A^i x} < a$, or to the value $b$ if $\boldsymbol{A^i x} > b$. This is called the `saturation effect', and is common in all sensing systems (not only the compressive ones). 

\textbf{Problem statement:} In this paper, we consider the following forward model for the measurements $\boldsymbol{y}$ for a compressive device with dynamic range $[-\tau,\tau]$:
\begin{equation}
\forall i \in \{1,2,...,m\}, y_i = \mathcal{C}(\boldsymbol{A^i x} + \eta_i;-\tau,\tau).
\label{eq:basic}
\end{equation}
Here the noise values are i.i.d., with $\eta_i \sim \mathcal{N}(0,\sigma^2)$ with known $\sigma$. Also $\mathcal{C}(q;a,b)$ is a saturation operator defined as follows:
\begin{eqnarray}
\mathcal{C}(q;a,b) = 
\begin{cases}
a \textrm{ if } q < a, \\ 
b \textrm{ if } q > b, \\ 
q \textrm{ if } q \in [a,b].
\end{cases}
\end{eqnarray}
Here $q < a$ is called `negative saturation' and $q > b$ is called `positive saturation'. Given this forward model with known $\boldsymbol{A}$ and $\tau$, we seek to recover a sparse/weakly-sparse vector $\boldsymbol{x}$ from its compressive measurements $\boldsymbol{y}$. 

\subsection{Previous Work} 
\label{subsec:previous_work}
There exists a moderate-sized literature on the problem of CS recovery from saturated measurements, which we summarize here. Right through this paper, we use $S^{-}, S^{+}$ to denote the sets that respectively consist of indices of negatively and positively saturated measurements, $S$ is the set of indices of all measurements, and the set of indices of non-saturated measurements is $S_{ns}\triangleq S-S^+-S^-$. The work in \cite{Laska2011} proposes two types of estimators for CS recovery from measurements with saturation effects and \emph{uniform quantization} (i.e. bounded) noise: (1) `saturation rejection' (\textsc{SR}), which weeds out saturated measurements and performs recovery only from the non-saturated measurements via the estimator: $\textrm{min} \|\boldsymbol{x}\|_1 \textrm{ s. t. } \sum_{i \in S_{ns}} (y_i - \boldsymbol{A^i x})^2 \leq \epsilon^2_{ns}$; and (2) `saturation consistency' (\textsc{SC}), which imposes the added constraint in the \textsc{SR} estimator, that the recovered signal $\boldsymbol{\hat{x}}$ should obey the conditions that $\forall i \in S^{-}, \boldsymbol{A^i \hat{x}} \leq -(\tau-\Delta)$ and $\forall i \in S^{+}, \boldsymbol{A^i \hat{x}} \geq \tau-\Delta$, where $\Delta$ denotes quantization width. The \textsc{SR} method potentially ignores many useful measurements (depending on the relation between $\tau$ and $\|\boldsymbol{x}\|_2$), and in the worst case the remaining part of the sensing matrix may not obey the RIP due to an insufficient number of measurements. The \textsc{SC} method is hard to adapt to saturation effects with \emph{Gaussian noise}, which is unbounded in nature. The work in \cite{Laska2009,Li2010} seeks to optimize the following cost function, which is based on the assumption that saturated measurements are not too large in number:
\begin{eqnarray}
J_{ss}(\boldsymbol{x}) \triangleq \lambda (\|\boldsymbol{x}\|_1 + \|\boldsymbol{r}\|_1) + \|\boldsymbol{y}-(\boldsymbol{A x}+\boldsymbol{r})\|^2_2 \nonumber \\
= \lambda \|\boldsymbol{x} ; \boldsymbol{r}\|_1 + \|\boldsymbol{y}-[\boldsymbol{A} | \boldsymbol{I}](\boldsymbol{x}; \boldsymbol{r}) \|^2_2.
\label{eq:SS}
\end{eqnarray}
Here $\boldsymbol{r}$ refers to the error due to saturation effects, $(\boldsymbol{x} ; \boldsymbol{r})$ is the concatenation of column vectors $\boldsymbol{x}, \boldsymbol{r}$; $\boldsymbol{I}$ is the identity matrix; and the $\|\boldsymbol{r}\|_1$ term promotes sparsity on the vector $\boldsymbol{r}$. In this paper, we term this approach `saturation sparsity' (\textsc{SS}). Although \cite{Laska2009,Li2010} prove RIP of $[\boldsymbol{A} | \boldsymbol{I}]$, that property is true only in an asymptotic sense as $m \rightarrow \infty$ (with $n \rightarrow \infty$ and $m/n \rightarrow 0$). In the realistic regime when $m$ is small, we have observed that such a technique has a tendency to estimate $\boldsymbol{r}$ to be a vector of all zeroes, due to the penalty on $\|\boldsymbol{r}\|_1$. Recent work in \cite{Tzagkarakis2019} proposes a greedy approximation algorithm to minimize the following cost function, designed to be resilient to measurement outliers:
\begin{equation}
J_{\alpha}(\boldsymbol{x}) \triangleq \|\boldsymbol{y}-\boldsymbol{A x}\|^p_p + \lambda \|\boldsymbol{x}\|_0; 0 < p < 1.
\end{equation}
An approximation algorithm to minimize such a cost function is essential, as the $\|\|_0$ pseudo-norm otherwise renders this problem to be NP-hard. Note that the approaches in \cite{Laska2009,Li2010,Tzagkarakis2019} were designed for \emph{general impulse noise} and \emph{not} for saturation effects, and hence these methods do not use knowledge of the saturation threshold $\tau$. Very recent work in \cite{Foucart2018} provides theoretical bounds for the following interesting estimator, termed `noise-cognizant $\ell_1$-minimization' (\textsc{NCLM}):
\begin{eqnarray}
\textrm{argmin}_{\boldsymbol{x},\boldsymbol{r}} \|\boldsymbol{x}\|_1 \textrm{ such that } (i) \mathcal{C}(\boldsymbol{A x} + \boldsymbol{r};-\tau,\tau) = \boldsymbol{y}, \\\nonumber (ii) \|\boldsymbol{r}\|_2 \leq \gamma \epsilon; (iii) \|\boldsymbol{x}\|_2 \leq \gamma' \mu \sqrt{m}.
\end{eqnarray}
The parameters $\gamma, \gamma', \mu$ need to be selected based on properties of the sensing matrix, $\epsilon$ is a bound on $\|\boldsymbol{y}-\boldsymbol{A x}\|_2$, and the vector $\boldsymbol{r}$ plays the same role as in Eqn. \ref{eq:SS}. Our method presented in this paper does not require the choice of so many parameters, nor does it require an upper bound on $\|\boldsymbol{x}\|_2$.\\ The rest of this paper is organized as follows. The main objective function and its properties are presented in Sec. \ref{sec:main_method}. Several numerical results are presented and discussed in Sec. \ref{sec:results}. We conclude in Sec. \ref{sec:concl} with a discussion of avenues for future work.

\section{Main Method}
\label{sec:main_method}
In this section, we first present the cost function which we seek to optimize, for CS recovery under saturated measurements. Although we consider the signal $\boldsymbol{x}$ to be sparse in the canonical basis, our method is easily extensible to a signal that in sparse/weakly sparse in any known orthonormal basis  (see Sec. \ref{sec:results}). In the following,  $\Phi(.)$ denotes the cumulative distribution function (CDF) of a standard normal random variable, and $\phi(.)$ denotes its probability density function (PDF). 
\subsection{Cost function and its properties}
Our cost function $J_{our}(\boldsymbol{x})$ is given below:
\begin{equation}
J_{our}(\boldsymbol{x}) = \lambda \|\boldsymbol{x}\|_1 + L(\boldsymbol{y},\boldsymbol{Ax};\tau),
\label{eqn:cf_our}
\end{equation}
where
\begin{eqnarray*}
L(\boldsymbol{y},\boldsymbol{Ax};\tau) \triangleq \dfrac{1}{2}\sum_{i \in S_{ns}} \Big(\dfrac{y_i - \boldsymbol{A^i x}}{\sigma}\Big)^2 \\ - \sum_{i \in S^+} \log\Big(1-\Phi((\tau-\boldsymbol{A^i x})/\sigma)\Big) - \sum_{i \in S^-} \log\Big(\Phi((-\tau-\boldsymbol{A^i x})/\sigma)\Big).
\end{eqnarray*}
The first term in $L(\boldsymbol{y},\boldsymbol{Ax};\tau)$ is due to the Gaussian noise in the unsaturated measurements; the second (third) term encourages the values of $\boldsymbol{A^i x}$, i.e. the members of $S^+$ (likewise $S^-$) to be much greater than $\tau$ (likewise much less than $-\tau$). To understand the behaviour of the second term of $L(\boldsymbol{y},\boldsymbol{Ax};\tau)$, consider a measurement $y_i$ such that $i \in S^+$. Referring to Eqn. \ref{eq:basic}, we have $P(y_i \geq \tau) = P(\eta_i \geq \tau-\boldsymbol{A^i x}) = 1-\Phi((\tau-\boldsymbol{A^i x})/\sigma)$. The last equality is due to the Gaussian nature of $\eta_i$. Given such a measurement, we seek to find $\boldsymbol{x}$ such that $\boldsymbol{A^i x} > \tau$, which will push $\tau-\boldsymbol{A^i x}$ toward $-\infty$, i.e. push $\Phi((\tau-\boldsymbol{A^i x})/\sigma)$ toward 0, and thus reduce the cost function. A similar argument can be made for the third term involving $S^-$. Consider that $P(y_i < -\tau) = P(\eta_i < -\tau - \boldsymbol{A^i x}) = \Phi((-\tau-\boldsymbol{A^ix})/\sigma)$. We seek to find $\boldsymbol{x}$, which will tend to push $-\tau-\boldsymbol{A^i x}$ toward $+\infty$, i.e. push $\Phi((-\tau-\boldsymbol{A^i x})/\sigma)$ toward 1, and thereby reduce the cost function. Assuming independence of the measurements, note that $L(\boldsymbol{y},\boldsymbol{Ax};\tau)$ is essentially the negative log of the following likelihood function:
\begin{flalign}
\tilde{L}(\boldsymbol{y},\boldsymbol{Ax};\tau) \triangleq \prod_{i \in S_{ns}} \dfrac{e^{-(y_i-\boldsymbol{A^ix})^2/(2\sigma^2)}}{\sigma\sqrt{2\pi}} \\\nonumber \prod_{i \in S^+}[1-\Phi((\tau-\boldsymbol{A^i x})/\sigma)]  \prod_{i \in S^-}\Phi((-\tau-\boldsymbol{A^i x})/\sigma).
\end{flalign}
We henceforth term our technique `likelihood maximization' or \textsc{LM}. The tendency to push $\Phi((\tau-\boldsymbol{A^i x})/\sigma)$ toward 0 or to push $\Phi((-\tau-\boldsymbol{A^i x})/\sigma)$ toward 1, is counter-balanced by the sparsity-promoting term $\|\boldsymbol{x}\|_1$, with $\lambda$ deciding the relative weightage.  $\blacksquare$ 
\vspace{-0.25in}
\subsection{Theoretical Analysis}
We now state an important property of $L(\boldsymbol{y},\boldsymbol{Ax};\tau)$, proved in the supplemental material \cite{suppmat}. \\
\textbf{Theorem 1:} $L(\boldsymbol{y},\boldsymbol{Ax};\tau)$ is a convex function of $\boldsymbol{x}$.$\blacksquare$ \\
For further theoretical analysis, we present an overview of the broad framework in \cite{Negahban2012} and then adapt it meticulously for the analysis of our estimator in Eqn. \ref{eqn:cf_our}. At first we state \textbf{L1, D1 and T1} and then we use them to prove \textbf{Theorems 2,3,4}.\\
\textbf{Lemma L1:} (Lemma 1 of \cite{Negahban2012}): Let $\widehat{\boldsymbol{x_{\lambda}}}$ be the optimum of a general cost function $L^g(\boldsymbol{y};\boldsymbol{Ax}) + \lambda \|\boldsymbol{x}\|_1$ with a regularization parameter $\lambda \geq 2 \|\nabla L^g(\boldsymbol{y};\boldsymbol{Ax})\|_{\infty}$. Then the error vector $\boldsymbol{\Delta} \triangleq \widehat{\boldsymbol{x_{\lambda}}}-\boldsymbol{x}$ belongs to the set $\mathbb{C}(S;\boldsymbol{x}) \triangleq \{\boldsymbol{\Delta}| \|(\boldsymbol{x}-\widehat{\boldsymbol{x_{\lambda}}})_{S^c}\|_1 \leq 3 \|(\boldsymbol{x}-\widehat{\boldsymbol{x_{\lambda}}})_S\|_1$, where $S$ is the set of indices of the $s$ non-zero elements of $\boldsymbol{x}$, and $\forall i \in S, x_S(i) = x_i;\forall i \notin S, x_S(i) = 0$. $\blacksquare$ \\ 
\textbf{Definition D1:} A loss function $L$ is said to obey the restricted strong convexity (RSC) property with curvature $\kappa_L > 0$ and tolerance function $\tau_L(\boldsymbol{x})$ if the Bregman divergence $\delta L^g(\boldsymbol{\Delta},\boldsymbol{x}) \triangleq L^g(\boldsymbol{y};\boldsymbol{A\widehat{\boldsymbol{x_{\lambda}}}}) - L^g(\boldsymbol{y};\boldsymbol{Ax}) - \nabla L^g(\boldsymbol{y};\boldsymbol{Ax})^t (\boldsymbol{\Delta})$ (the error between the loss function value at $\widehat{\boldsymbol{x_{\lambda}}}$ and its first order Taylor series expansion about $\boldsymbol{x}$) satisfies
$\delta L^g(\boldsymbol{\Delta},\boldsymbol{x}) \geq \kappa_L \|\boldsymbol{\Delta}\|^2_2 - \tau^2_L(\boldsymbol{x})$ for every vector $\boldsymbol{\Delta} \in \mathbb{C}(S;\boldsymbol{x})$. $\blacksquare$ \\
Intuitively, a loss function that obeys RSC is sharply curved around $\boldsymbol{x}$, so that any difference in the loss function $|L^g(\boldsymbol{y};\boldsymbol{Ax})-L^g(\boldsymbol{y};\boldsymbol{A\widehat{\boldsymbol{x_{\lambda}}}})|$ will imply a \emph{proportional} estimation error $\|\boldsymbol{x}-\widehat{\boldsymbol{x_{\lambda}}}\|_1$ for all error vectors $\widehat{\boldsymbol{x_{\lambda}}}-\boldsymbol{x} \in \mathbb{C}(S;\boldsymbol{x})$. We refer the reader to \cite{Negahban2012} for more details. \\
\textbf{Theorem T1:} (Theorem 1 of \cite{Negahban2012}) If $L^g$ is convex, differentiable and obeys RSC property with curvature $\kappa_L$ and tolerance $\tau^2_L(\boldsymbol{x})$, if $\widehat{\boldsymbol{x_{\lambda}}}$ is as defined in Lemma L1 with $\lambda \geq 2 \|\nabla L(\boldsymbol{y};\boldsymbol{Ax})\|_{\infty}$, and if $\boldsymbol{x}$ is an $s$-sparse vector, then we have:
$\|\widehat{\boldsymbol{x_{\lambda}}}-\boldsymbol{x}\|^2_2 \leq \dfrac{9 \lambda^2 s}{\kappa^2_L} + \dfrac{2\lambda \tau^2_L(\boldsymbol{x})}{\kappa_L}$.$\blacksquare$\\
We now state the following theorems pertaining to the cost function in Eqn. \ref{eqn:cf_our} and prove them in \cite{suppmat}:\\
\textbf{Theorem 2:} $L(\boldsymbol{y},\boldsymbol{Ax};\tau)$ from Eqn. \ref{eqn:cf_our} follows RSC with curvature $\kappa_L=\frac{\gamma}{2\sigma^2}$ and tolerance function $\tau_{L}^2(\boldsymbol{x}) = 0$, where $\gamma$ is the restricted eigenvalue constant (REC) for $\boldsymbol{A}$.\\ Here, we use the structure of $\delta L^g(\boldsymbol{\Delta},\boldsymbol{x})$ defined in \textbf{D1} to find the values of curvature and tolerance function for our cost function. Proving RSC for our cost function implies that we will reach the global minima. $\blacksquare$\\
\textbf{Theorem 3:} For our noise model and with additional constraints on the signal that $\forall i, \alpha \leq x_i \leq \beta$, we have the lower bound $\|\nabla L\|_{\infty} \geq \frac{\sqrt{\varrho\log(n)}}{\sigma\sqrt{m}}\{\sqrt{{m_3}}+C_1 \sqrt{{(m_1+m_2)}}\}$ with probability $1-2\exp{-\frac{1}{2}(\varrho-2)\log(n)}$ for constant $C_1, \varrho > 2$.\\
We develop this lower bound for $\|\nabla L\|_{\infty}$ so that we can apply \textbf{T1} to find the upper bound on the reconstruction error in \textbf{Theorem 4} $\blacksquare$\\
\textbf{Theorem 4:} Let $\widehat{\boldsymbol{x}_{\lambda}}$ be the minimizer of the cost function in Eqn. \ref{eqn:cf_our} with regularization parameter $\lambda \geq 2 \|\nabla L\|_{\infty}$ and with the signal constraints from Thm. 3. Let $\boldsymbol{x}$ be the true $s$-sparse signal which gave rise to the compressive measurements in $\boldsymbol{y}$. Then we have the following upper bound with the same probability as in Thm. 3:
$\|\widehat{\boldsymbol{x}_{\lambda}}-\boldsymbol{x}\|_2^2 \leq 
     \dfrac{144s\log(n)\sigma^2\varrho}{\gamma^2{m}} ( \sqrt{{m_3}}+
     C_1\sqrt{{(m_1+m_2)}})^2. \blacksquare$
\\
\textbf{Observations related to the upper bound:} 
The upper bound is directly proportional to $s\log(n)$ which is equivalent to the upper bound in Lasso reconstruction. So, the tightness of the upper bound on the reconstruction error of our cost function is relatively close to that of Lasso reconstruction.
The bound is directly proportional to $\sigma^2$ as well as $s = \|\boldsymbol{x}\|_0$ and inversely proportional to $\gamma = \textrm{REC}(\boldsymbol{A};s)$ \cite{Raskutti2010,THW2015}, all of which is very intuitive. The bound also becomes looser with increase in the number of saturated measurements $m_1,m_2$. If there are no saturated measurements, i.e. $m_1 = m_2 = 0$, then the bound reduces to the normal LASSO bound \cite{THW2015}, except that here we consider $\boldsymbol{A}$ with unit column norm as against column norm of $m$ in \cite{THW2015}. The bound also increases with $m_3$. However, it turns out that the constant factor $C_1$ for the $O(\sqrt{m_1+m_2})$ term in the bounds, is very large. This is because it contains other factors of the form $\frac{\phi(z)}{\Phi(z)}$ or $\frac{\phi(z)}{1-\Phi(z)}$ where $z$ stands for either $\alpha$ or $\beta$ (see suppl. mat. \cite{suppmat}), which are both large in absolute value for large $\alpha,\beta$. Hence the $O(\sqrt{m_1 + m_2})$ term dominates over the $O(\sqrt{m_3})$ term, which is intuitive. 

\section{Experimental Results}
\label{sec:results}
Here we report results on CS recovery using our technique \textsc{LM} in comparison to the following existing approaches described in Sec. \ref{subsec:previous_work}: (i) Saturation rejection (\textsc{SR}) from \cite{Laska2011}; (ii) Saturation Consistency (\textsc{SC}) from \cite{Laska2011} with the following constraint set designed to handle Gaussian measurement noise: $\forall i \in S^{-}, \boldsymbol{A^i \hat{x}} \leq -\tau + 3\sigma$ and $\forall i \in S^{+}, \boldsymbol{A^i \hat{x}} \geq \tau-3\sigma$; (iii) Saturation Sparsity (\textsc{SS}) from \cite{Li2010}, (iv) Saturation Ignorance (\textsc{SI}), a technique which recovers $\boldsymbol{x}$ pretending there was no saturation in $\boldsymbol{y}$; and (v) \textsc{NCLM} from \cite{Foucart2018}. Define $\zeta \triangleq \sum_{i=1}^m |\boldsymbol{A^i x}|/m$, the average absolute value of noiseless unsaturated measurements. For \emph{all techniques} including \textsc{LM}, we assume knowledge of $\tau$ and thereby that of sets $S^+,S^-$. For \textsc{LM}, we did not impose the constraints $\alpha \leq x_i \leq \beta$ from Thm. 3, due to negligible impact on the results.
\\
\textbf{Experiment description:} All our experiments were performed on signals of dimension $n = 256$ that were sparse in the 1D-DCT (discrete cosine transform) basis. The supports of the DCT coefficient vectors were chosen randomly, and each signal had a different support. The elements of the sensing matrix $\boldsymbol{A}$ were drawn i.i.d. from $\mathcal{N}(0,1/m)$ so that $\boldsymbol{A}$ would obey RIP with high probability \cite{Candes2008}. Gaussian noise was added to the measurements, followed by application of the saturation operator $\mathcal{C}$. Keeping all other parameters fixed, we studied the variation in the performance of these six techniques with regard to change in (\underline{A}) number of measurements $m$; (\underline{B}) signal sparsity $s$ expressed as fraction $f_{sp} \in [0,1]$ of signal dimension $n$; (\underline{C}) noise standard deviation $\sigma$ expressed as a fraction $f_{\sigma} \in [0,1]$ of $\zeta$; and (\underline{D}) the fraction $f_{sat} \in [0,1]$ of the $m$ measurements that were saturated. For the measurements experiment (i.e. (\underline{A})), $m$ was varied in $\{30,40,50,...,250\}$ with $s = 25, f_{sat} = 0.15, f_{\sigma} = 0.1$. For the sparsity experiment (i.e. (\underline{B})), $f_{sp}$ was varied in $\{0.05,0.1,0.15,0.2\}$ with $m = 150, f_{sat} = 0.15, f_{\sigma} = 0.1$. For the noise experiment (i.e. (\underline{C})), we varied  $f_{\sigma}$ in $\{0.01,0.02,0.04,...,0.2\}$ with $m=150,f_{sp}=25/256,f_{sat}=0.15$. For the saturation experiment (i.e. (\underline{D})), $f_s$ was varied in $\{0,5,10,...,50\}/150$ with $m=150,f_{sp}=25/256,f_{\sigma}=0.1$. The performance was measured using relative root-mean squared error (RRMSE) (defined as $\|\boldsymbol{x}-\boldsymbol{\hat{x}}\|_2/\|\boldsymbol{x}\|_2$ where $\boldsymbol{\hat{x}}$ is an estimate of the signal $\boldsymbol{x}$), computed over reconstructions from 10 noise trials.\\ 
\begin{figure}
    \begin{center}
    {
    \includegraphics[scale=0.2]{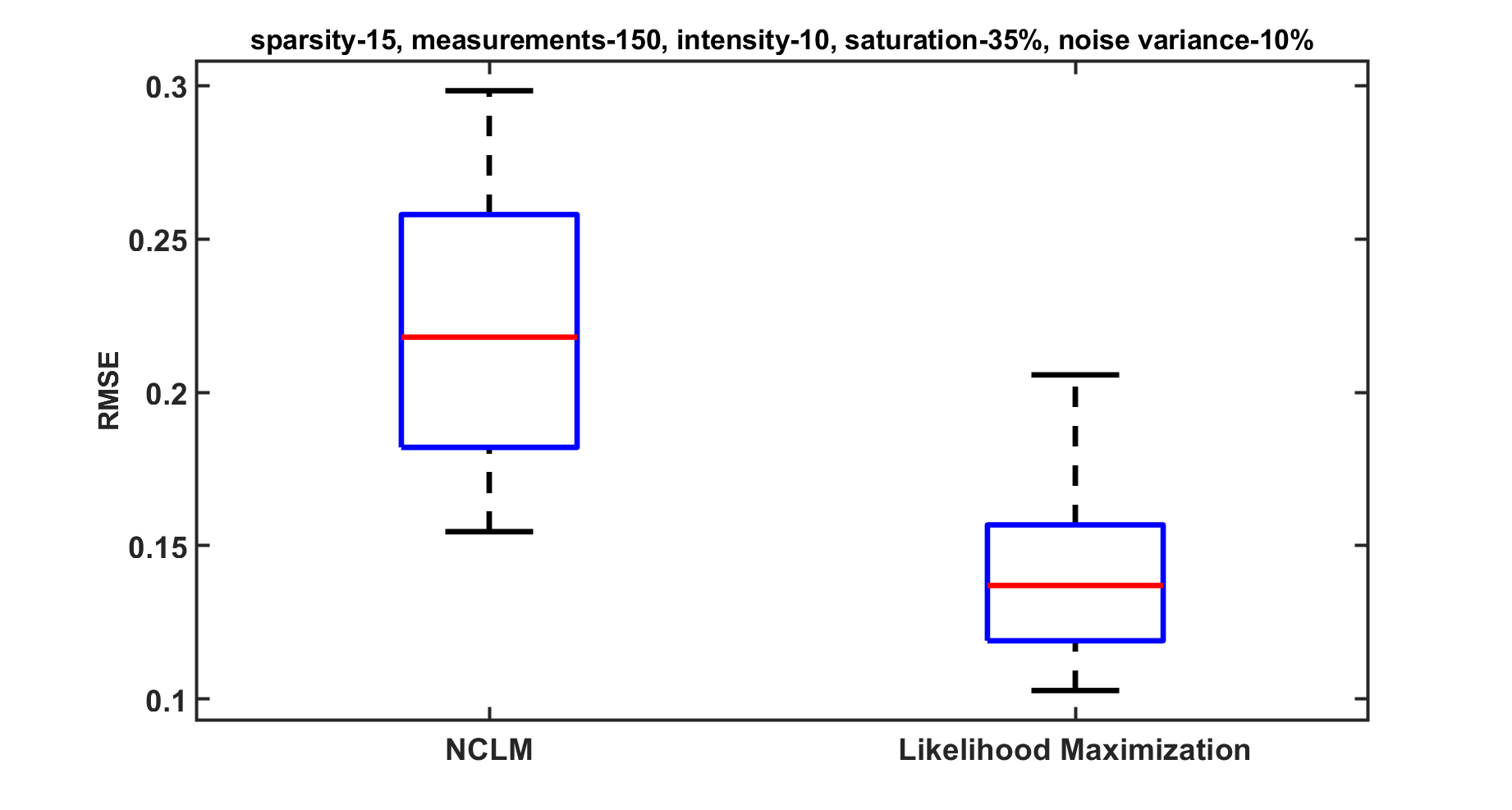}
    \caption{Comparison of \textsc{NCLM} and \textsc{LM} for $s = 15, m = 150, n = 256, f_{sat} = 0.35, f_{sig} = 0.1$. }
    \label{fig:boxplot_SR_LM}
    }
    \end{center}
\end{figure}
\begin{figure}
    \includegraphics[scale=0.19]{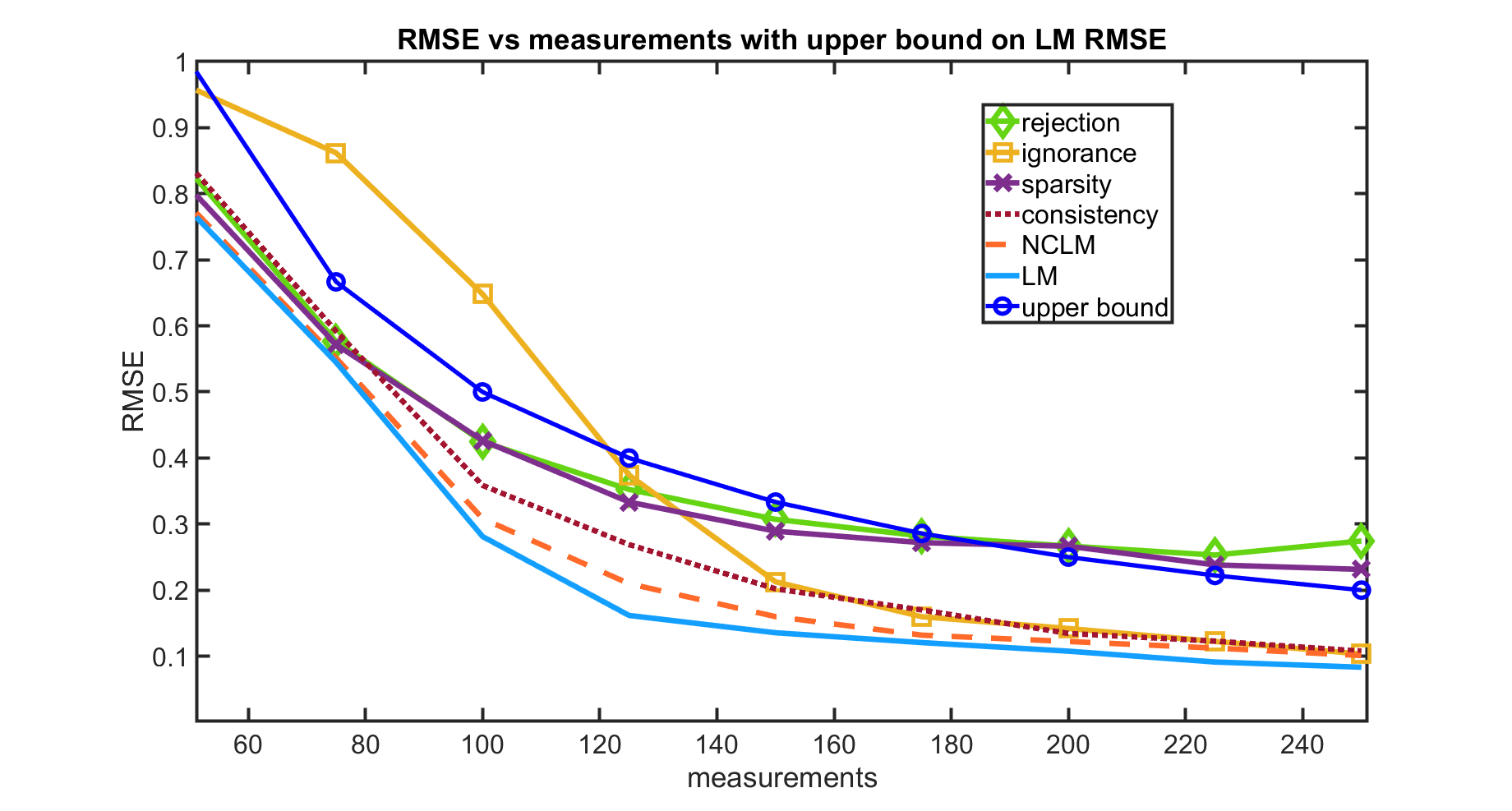}
    \includegraphics[scale=0.19]{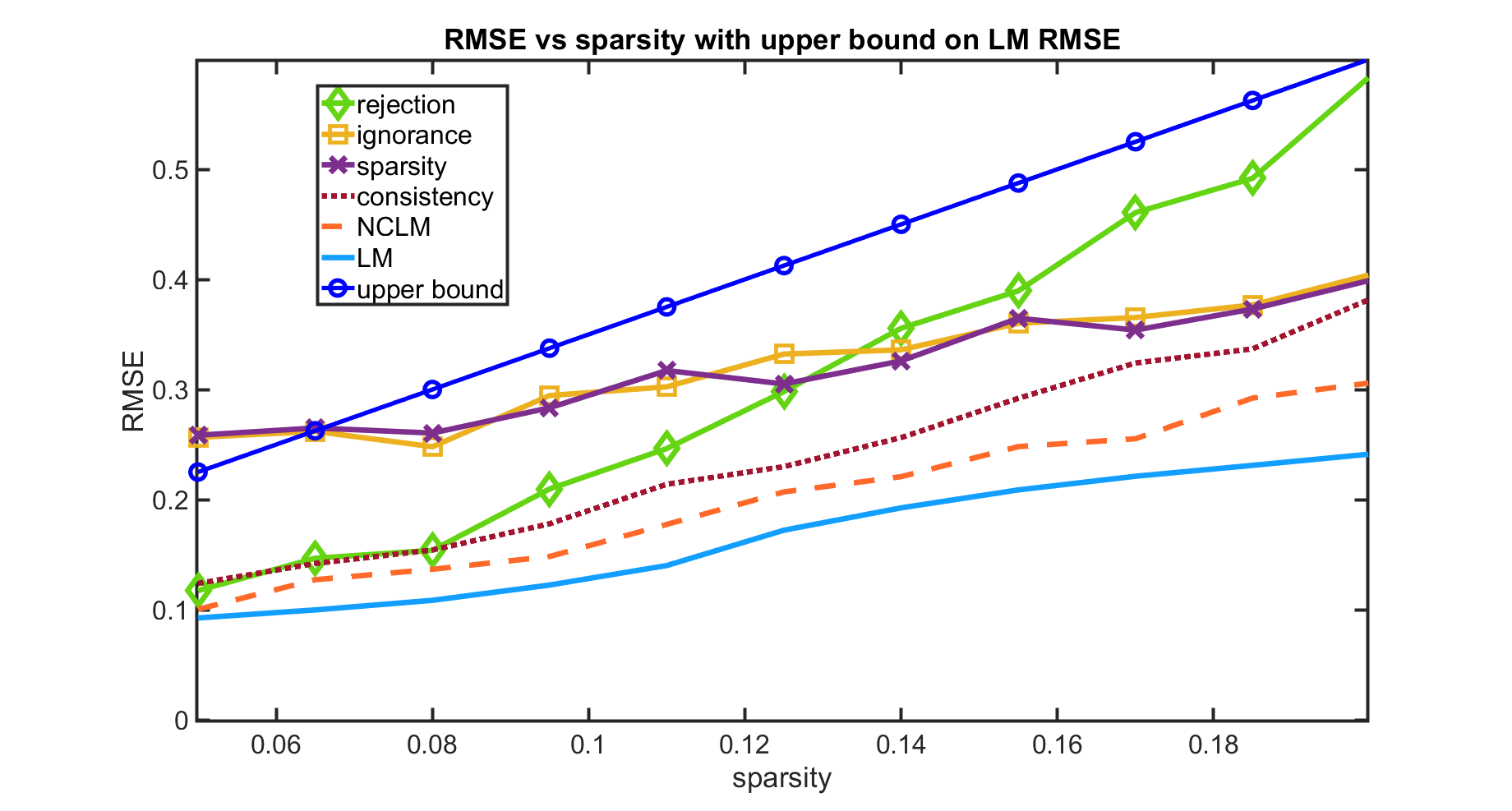}
    \includegraphics[scale=0.19]{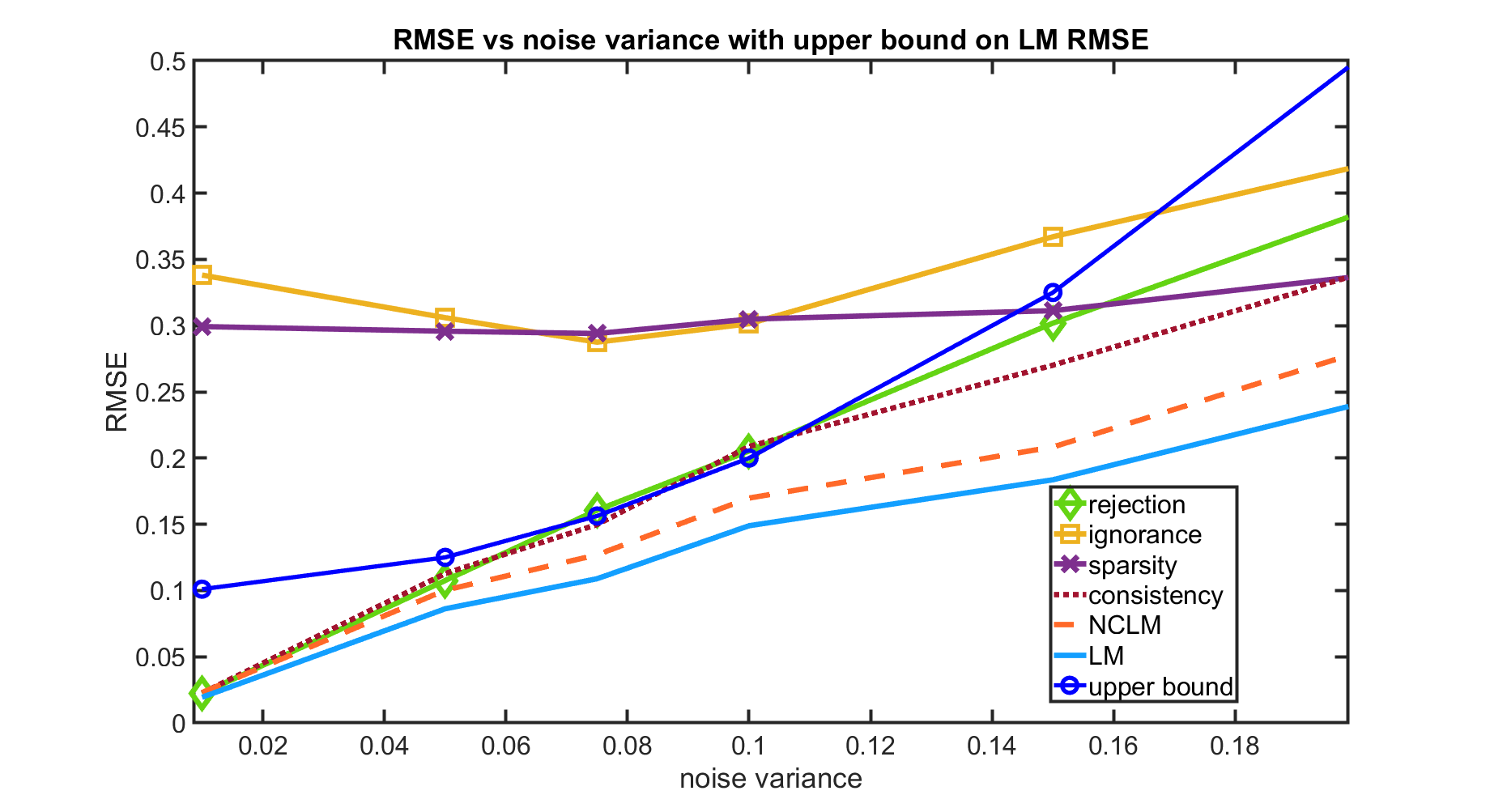}
    \includegraphics[scale=0.19]{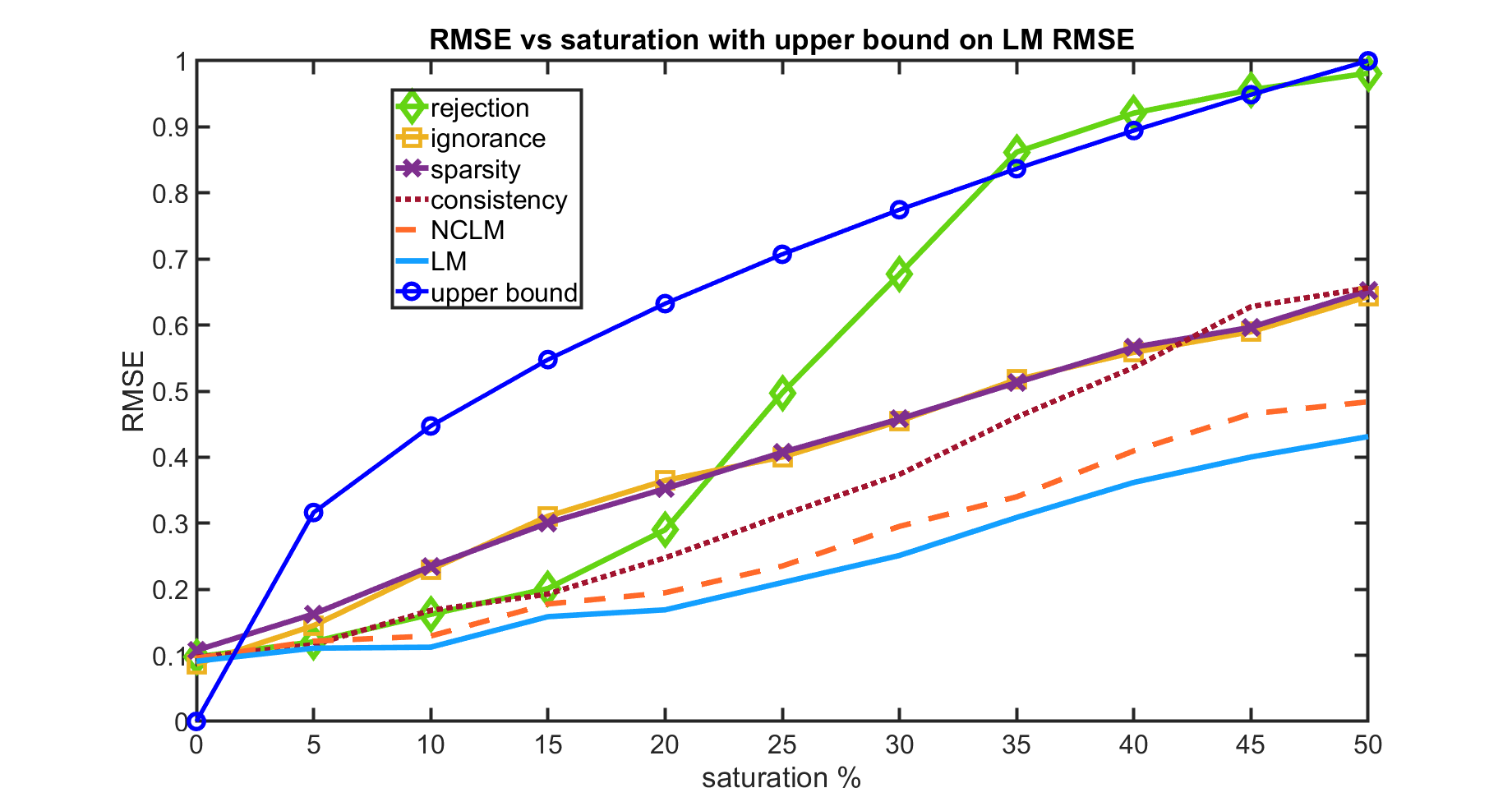}
    \caption{Performance comparison for six methods: \textsc{SR} (saturation rejection), \textsc{SC} (saturation consistency), \textsc{SI} (saturation ignorance), \textsc{SS} (saturation sparsity), the \textsc{NCLM} method and the proposed \textsc{LM} technique w.r.t. variation in number of measurements $m$ (topmost, experiment (\underline{A})), signal sparsity $s$ (2nd from top, experiment (\underline{B})), noise $\sigma$ (3rd from top, experiment (\underline{C})) and fraction $f_s$ of the $m$ measurements that were saturated (bottom-most, experiment (\underline{D})).}
    \label{fig:exp_results}    
\end{figure}
\textbf{Parameter settings:} For the proposed \textsc{LM} technique and for \textsc{SS}, the regularization parameter $\lambda$ was chosen using cross-validation on a set of unsaturated measurements, following the method in \cite{Zhang2014}. The size of the cross-validation set was 0.3 times the number of measurements used for reconstruction. For \textsc{SR} and \textsc{SC}, we set $\epsilon_{ns} = \sigma \sqrt{|S_{ns}|}$. For \textsc{SI}, we used the estimator $\textrm{min} \|\boldsymbol{x}\|_1 \textrm{ s. t. } \|\boldsymbol{y}-\boldsymbol{Ax}\|_2 \leq \sigma\sqrt{m}$. For \textsc{NCLM}, the bound on $\|\boldsymbol{x}\|_2$ was set to be the $\ell_2$-norm of the true signal (omnisciently), and that on $\|\boldsymbol{r}\|_2$ was set to be a statistical estimate of the magnitude of the pre-saturated noise vector. The well-known FISTA algorithm \cite{Beck2009} was used for \textsc{LM}, whereas CVX was used for \textsc{SS}, \textsc{SC}, \textsc{SR} and \textsc{SI}. \\
\textbf{Discussion:} The results of these experiments are summarized in Fig. \ref{fig:exp_results}, and show that the proposed \textsc{LM} technique consistently outperforms the competing methods numerically. This behaviour is particularly observable for high $f_{sat}$ or $f_{sig}$. We observed that \textsc{SC} outperformed \textsc{SR} for high $f_{sat}$ or $f_{sig}$. We also note that our technique performed better than \textsc{NCLM} (our closest competitor) in the regime of high $f_{sig}$ and high $f_{sat}$, as can be seen from Fig. \ref{fig:boxplot_SR_LM}.The upper bound of the reconstruction error is plotted using proper scaling . The empirical trends observed here clearly satisfies the intuitive arguments however, tightness of the bound might vary depending on the constants of the upper bound. 
\section{Conclusion}
\label{sec:concl}
We have presented a principled likelihood-based method of compressive signal recovery under Gaussian noise combined with saturation effects. We have proved the convexity of our estimator and derived the upper performance bound, and shown that it numerically outperforms competing methods. The recent work in \cite{Bohra2019} handles compressive inversion under with Poisson-Gaussian-uniform quantization noise, a very realistic noise model in imaging systems. Extending the numerical simulations as well as the convexity proofs to handle saturation effects in conjunction with such a Poisson-Gaussian noise model is a potential avenue for future work. Another useful avenue of research would be to derive lower performance bounds for the presented penalized estimator. 
\section{Appendix}

This section contains the proof of various results from the main paper. All theorem numbers refer to the corresponding ones in the main paper. 

\subsection{Cost function and its properties}
\label{sec:cost_fn}
Our cost function $J_{our}(\boldsymbol{x})$ is given below:\\
\begin{eqnarray*}
L(\boldsymbol{y},\boldsymbol{Ax};\tau) \triangleq \dfrac{1}{2}\sum_{i \in S_{ns}} \Big(\dfrac{y_i - \boldsymbol{A^i x}}{\sigma}\Big)^2 \\ - \sum_{i \in S^+} \log\Big(1-\Phi((\tau-\boldsymbol{A^i x})/\sigma)\Big) - \sum_{i \in S^-} \log\Big(\Phi((-\tau-\boldsymbol{A^i x})/\sigma)\Big),\\
J_{our}(\boldsymbol{x}) = \lambda \|\boldsymbol{x}\|_1 + L(\boldsymbol{y},\boldsymbol{Ax};\tau).
\label{eqn:cf_our}
\end{eqnarray*}\\
The notation $\boldsymbol{A^i}$ denotes the i-th row of the sensing matrix $\boldsymbol{x}$.The first term in $L(\boldsymbol{y},\boldsymbol{Ax};\tau)$ is due to the Gaussian noise in the unsaturated measurements; the second (third) term encourages the values of $\boldsymbol{A^i x}$, i.e. the members of $S^+$ (likewise $S^-$) to be much greater than $\tau$ (likewise much less than $-\tau$). To understand the behaviour of the second term of $L(\boldsymbol{y},\boldsymbol{Ax};\tau)$, consider a measurement $y_i$ such that $i \in S^+$. We have $P(y_i \geq \tau) = P(\eta_i \geq \tau-\boldsymbol{A^i x}) = 1-\Phi((\tau-\boldsymbol{A^i x})/\sigma)$. The last equality is due to the Gaussian nature of $\eta_i$. Given such a measurement, we seek to find $\boldsymbol{x}$ such that $\boldsymbol{A^i x} > \tau$, which will push $\tau-\boldsymbol{A^i x}$ toward $-\infty$, i.e. push $\Phi((\tau-\boldsymbol{A^i x})/\sigma)$ toward 0, and thus reduce the cost function. A similar argument can be made for the third term involving $S^-$. Consider that $P(y_i < -\tau) = P(\eta_i < -\tau - \boldsymbol{A^i x}) = \Phi((-\tau-\boldsymbol{A^ix})/\sigma)$. We seek to find $\boldsymbol{x}$, which will tend to push $-\tau-\boldsymbol{A^i x}$ toward $+\infty$, i.e. push $\Phi((-\tau-\boldsymbol{A^i x})/\sigma)$ toward 1, and thereby reduce the cost function. Assuming independence of the measurements, note that $L(\boldsymbol{y},\boldsymbol{Ax};\tau)$ is essentially the negative log of the following likelihood function:
\begin{flalign} \label{eq:1}
\tilde{L}(\boldsymbol{y},\boldsymbol{Ax};\tau) \triangleq \prod_{i \in S_{ns}} \dfrac{e^{-(y_i-\boldsymbol{A^ix})^2/(2\sigma^2)}}{\sigma\sqrt{2\pi}} \\\nonumber \prod_{i \in S^+}[1-\Phi((\tau-\boldsymbol{A^i x})/\sigma)]  \prod_{i \in S^-}\Phi((-\tau-\boldsymbol{A^i x})/\sigma).
\end{flalign}
We henceforth term our technique `likelihood maximization' or \textsf{LM}. The tendency to push $\Phi((\tau-\boldsymbol{A^i x})/\sigma)$ toward 0 or to push $\Phi((-\tau-\boldsymbol{A^i x})/\sigma)$ toward 1, is counter-balanced by the sparsity-promoting term $\|\boldsymbol{x}\|_1$, with $\lambda$ deciding the relative weightage. \\
Note that since we assume $\tau$ is known, we know the exact constitution of $S^+,S^-$ in a data-driven manner \emph{in all the techniques including ours}, i.e. we assign the $i^{\textrm{th}}$ measurement to $S^{+}$ if $y_i = \tau$ and to $S^-$ if $y_i = -\tau$. 
\subsection{Proof of Theorem 1: Convexity of Data Fidelity Term}
We now state and prove an important property of $L(\boldsymbol{y},\boldsymbol{Ax};\tau)$. \\
\textbf{Theorem 1:} $L(\boldsymbol{y},\boldsymbol{Ax};\tau)$ is a convex function of $\boldsymbol{x}$. $\blacksquare$ \\
\textbf{Proof:} For proving convexity, we show that the Hessian matrix $\dfrac{\partial^2 L}{\partial \boldsymbol{x} \partial \boldsymbol{x}^t}$ is positive semi-definite. Define $Q_1(\boldsymbol{x}) \triangleq \dfrac{1}{2}\sum_{i \in S_{ns}} \Big(\dfrac{y_i - \boldsymbol{A^i x}}{\sigma}\Big)^2; Q_2(\boldsymbol{x}) \triangleq - \sum_{i \in S^+} \log\Big(1-\Phi((\tau-\boldsymbol{A^i x})/\sigma)\Big); Q_3(\boldsymbol{x}) \triangleq - \sum_{i \in S^-} \log\Big(\Phi((-\tau-\boldsymbol{A^i x})/\sigma)\Big)$. It is clear that $\dfrac{\partial^2 Q_1}{\partial \boldsymbol{x} \partial \boldsymbol{x}^t} = \sum_{i \in S_{ns}} \boldsymbol{A^i}(\boldsymbol{A^i})^t$, which is a positive semi-definite matrix. Now, we have: \\
\begin{eqnarray*}
    \dfrac{\partial Q_2(\boldsymbol{x})}{\partial \boldsymbol{x}}
    =-\dfrac{1}{\sigma} \sum_{i \in S^+}{\Big(\boldsymbol{A^i}\phi(\dfrac{\tau-\boldsymbol{A^i x}}{\sigma})\Big)/\Big(1-\Phi(\dfrac{\tau-\boldsymbol{A^i x}}{\sigma})}\Big),\\
    \dfrac{\partial^2 Q_2(\boldsymbol{x})}{\partial \boldsymbol{x} \partial \boldsymbol{x}^t}
    =\dfrac{1}{\sigma^2}\sum_{i \in S^+}\boldsymbol{A^i} \boldsymbol{A^i}^t h_i, 
    \end{eqnarray*}
    where
    \begin{equation}\label{eq:2}
    h_i \triangleq \dfrac{(\phi(u_i))^2-[1-\Phi(u_i)]u_i\phi(u_i)}{\Big[1-\Phi(u_i)\Big]^2}; u_i \triangleq \dfrac{\tau-\boldsymbol{A^i x}}{\sigma}.
    \end{equation}
In the expression for the Hessian of $Q_2$, we note that terms such as $\boldsymbol{A^i}\boldsymbol{A^i}^t$ form a positive semi-definite matrix $\forall i$, and the denominator in every $h_i$ is non-negative. If we can prove that the numerator of each term $h_i$ is non-negative as well, then we can show $Q_2$ to be convex since its Hessian would be positive semi-definite. The numerator has the form
$\phi(u) H(u)$ where $H(u) \triangleq \phi(u)-[1-\Phi(u)]u$. Since $\phi(u) \geq 0$ always, we just have to prove that $H(u) \geq 0$. We see that $H(0)=1/\sqrt{2\pi}$ ; $H(\infty)=\lim_{u \rightarrow \infty}{\phi(u)-[1-\Phi(u)]u} = 0$. The latter is because as $u\rightarrow \infty$ , $\phi(u) \rightarrow 0$ , $[1-\Phi(u)] \rightarrow 0$. But the rate of convergence of $[1-\Phi(u)] \rightarrow 0$ is faster than that of $u\rightarrow \infty$ on the extended real line, so $H(\infty)=0$. Also $H(-\infty) = \infty$. Noting that $\phi^{'}(u)=-u \phi(u)$, we see that $H^{'}(u) = \phi^{'}(u)-1+u \phi(u) + \Phi(u) = \Phi(u) - 1 \leq 0$. Hence $H(u$) is a non-increasing function bounded below by 0, which establishes that $H(u) \geq 0$ for all $u \in \mathbb{R}$, and hence $\forall i, h_i \geq 0$. 
Since $\phi(u) \geq 0$, we see that $\phi(u) H(u) \geq 0$. This establishes that $Q_2$ is convex. We can establish the convexity of $Q_3$ along very similar lines. 
\begin{eqnarray*}
\dfrac{\partial^2 Q_3(\boldsymbol{x})}{\partial \boldsymbol{x} \partial \boldsymbol{x}^t}
=\dfrac{1}{\sigma^2}\sum_{i \in S^{-}}\boldsymbol{A^i}\boldsymbol{A^i}^t g_i
\end{eqnarray*}
where $g_i \triangleq \dfrac{(\phi(u_i))^2+[\Phi(u_i)](u_i\phi(u_i))}{[\Phi(u_i)]^2}; u_i \triangleq \dfrac{\tau-\boldsymbol{A^i x}}{\sigma}.$
Using the same technique as for $Q_2$, we can show that $Q_3$ is convex. Since $Q_1, Q_2, Q_3$ are all convex, the convexity of $L$ follows. $\blacksquare$

\subsection{Proof of Theorem 2: Restricted Strong Convexity of $L(\boldsymbol{y},\boldsymbol{Ax};\tau)$}
The restricted Strong Convexity (RSC) is an important property for a data fidelity function from the point of proving performance bounds. Let $\Delta \triangleq \widehat{\boldsymbol{x}_{\lambda}}-\boldsymbol{x}^*$ be the difference between an optimal solution and the true parameter $\boldsymbol{x}^*$, and consider the loss difference $\mathcal{L}(\widehat{\boldsymbol{x}_{\lambda}}) - \mathcal{L}({\boldsymbol{x}^*})$ as defined in \cite{Negahban2012} (Lemma-1). In the classical setting, it is expected that the loss difference goes to zero with increase in sample size of the data in the model. However, convergence in the loss difference is not sufficient to imply $\Delta$ is small. This sufficiency depends on the curvature of the loss function. If the loss function is `sharply curved' around the optimal value $\widehat{\boldsymbol{x}_{\lambda}}$, then having a small loss difference implies having a small $\Delta$. However, if the loss function is `relatively constant' around the optimal value $\widehat{\boldsymbol{x}_{\lambda}}$, then the loss difference may be small but $\Delta$ can be relatively large. If the loss function is `too flat' around the optimal value, it may hamper the convergence of the optimal solution to the true value. Thus, to ensure that the loss function is not `too flat', the notion of strong convexity is considered.  One way to enforce that our cost function $\mathcal{L}$ is strongly convex is to require the existence of some positive constant $\kappa_L$ such that $\delta L(\boldsymbol{x}^*,\boldsymbol{\Delta}) \geq \kappa_L$ where $\delta L(\boldsymbol{x}^*,\boldsymbol{\Delta}) = L(\boldsymbol{x}^*+\boldsymbol{\Delta}) - L(\boldsymbol{x}^*)-<\boldsymbol{\Delta},\nabla L(\boldsymbol{x}^*)> \quad \forall \Delta \in \mathbb{R}^n$. This ensures that our optimal solution, if it exists, will reach the unique global minimum at a linear convergence rate. 
However strong convexity is impossible for all vectors in $\mathbb{R}^n$ in our case, as the matrix $\boldsymbol{A}$ is low rank (size $m \times n, m < n$). Instead, the notion of strong convexity defined over a restricted space of $\Delta$ is Restricted Strong Convexity (RSC) defined as follows (see Lemma 1 of \cite{Negahban2012}):
\begin{equation} \label{eq:3}
    \delta L(\boldsymbol{x^*},\boldsymbol{\Delta}) \geq \kappa_L \|\boldsymbol{\Delta}\|^2 - \tau_L^2(\boldsymbol{x^*}) \\
\end{equation}
for the curvature term $\kappa_L > 0$ and a positive tolerance function $\tau_L$ for, $\boldsymbol{\Delta} \in C$ such that $C \triangleq \{\boldsymbol{\Delta} :\|\boldsymbol{{\Delta}_{S^c}}\|_1 \leq 3\|\boldsymbol{\Delta_S}\|_1+4\|\boldsymbol{x^*_{S^c}}\|_1 \}$ as defined in \cite{Negahban2012}. Here $S$ stands for the set of indices of the $s$ largest entries of the true signal $\boldsymbol{x^*}$, and $S^c$ is the complement of $S$. This paper primarily considers purely sparse signals, and hence $S$ would correspond to the $s$ non-zero entries of $\boldsymbol{x^*}$ due to which $\boldsymbol{x^*_{S^c}} = \boldsymbol{0}$, but extensions to weakly sparse signals are also easily possible. 

\subsubsection{The form of the function $\delta L(\boldsymbol{x^*},\boldsymbol{\Delta})$}
Define,
\begin{multline}
 {A}(\boldsymbol{x^*},\boldsymbol{\Delta})  = L(\boldsymbol{y},\boldsymbol{A}(\boldsymbol{x^*}+\boldsymbol{\Delta});\tau)-L(\boldsymbol{y},\boldsymbol{Ax^*};\tau)  \\
  =-\sum^{m_1}_{i=1}{\ln{[1-\Phi(\frac{\tau-{A}^{i}(\boldsymbol{x^*}+\boldsymbol{\Delta})}{\sigma})]}} -\sum^{m_2}_{i=m_1+1}{\ln{[\Phi(\frac{-\tau-\boldsymbol{A}^i(\boldsymbol{x^*}+\boldsymbol{\Delta})}{\sigma})]}} \\ +\frac{1}{2}\sum^{m_3}_{i=m_2+1}{(\frac{y_i-\boldsymbol{A}^i(\boldsymbol{x^*}+\boldsymbol{\Delta})}{\sigma})^2}+\sum^{m_1}_{i=1}{\ln{[1-\Phi(\frac{\tau-\boldsymbol{A}^{i}\boldsymbol{x}}{\sigma})]}} +\sum^{m_2}_{i=m_1+1}{\ln{[\Phi(\frac{-\tau-\boldsymbol{A}^i\boldsymbol{x^*}}{\sigma})]}} -\frac{1}{2}\sum^{m_3}_{i=m_2+1}{(\frac{y_i-\boldsymbol{A}^i\boldsymbol{x^*}}{\sigma})^2} \\
   = -\sum^{m_1}_{i=1}({\ln{[1-\Phi(\frac{\tau-\boldsymbol{A}^{i}(\boldsymbol{x^*}+\boldsymbol{\Delta})}{\sigma})]}}-{\ln{[1-\Phi(\frac{\tau-\boldsymbol{A}^{i}\boldsymbol{x^*}}{\sigma})]}}) \\ -\sum^{m_2}_{i=m_1+1}({\ln{[\Phi(\frac{-\tau-\boldsymbol{A}^i(\boldsymbol{x^*}+\boldsymbol{\Delta})}{\sigma})]}}-{\ln{[\Phi(\frac{-\tau-\boldsymbol{A}^i\boldsymbol{x^*}}{\sigma})]}})
   \\+\frac{1}{2}\sum^{m_3}_{i=m_2+1}({(\frac{y_i-\boldsymbol{A}^i(\boldsymbol{x^*}+\boldsymbol{\Delta})}{\sigma})^2}-{(\frac{y_i-\boldsymbol{A}^i\boldsymbol{x}}{\sigma})^2}).
\end{multline}
\\ \\
In $\mathcal{A}(\boldsymbol{x^*},\boldsymbol{\Delta})$ , we have 3 terms: \\
Simplifying the $3^{rd}$  term, \\
\begin{multline}
{\dfrac{1}{2}}{\sum^{m_3}_{i=m_2+1}({(\frac{y_i-\boldsymbol{A}^i(\boldsymbol{x^*}+\boldsymbol{\Delta})}{\sigma})^2}-{(\frac{y_i-\boldsymbol{A}^i\boldsymbol{x}}{\sigma})^2})}\\
= {\dfrac{1}{2\sigma^2}}{\sum^{m_3}_{i=m_2+1}{\{y_i^2+[\boldsymbol{A^i}(\boldsymbol{x^*}+\boldsymbol{\Delta})]-2y_{i}\boldsymbol{A^i(x^*+\Delta)}-y_i^2-\{\boldsymbol{A^ix^*}\}^2+2y_{i}\boldsymbol{A^ix^*}\}}}\\ \nonumber
={\dfrac{1}{2\sigma^2}}{\sum^{m_3}_{i=m_2+1}{\{(\boldsymbol{A}^i\boldsymbol{x^*})^{T}(\boldsymbol{A}^i\boldsymbol{x^*})+2\boldsymbol{x^*}^T{\boldsymbol{A}^i}^T\boldsymbol{A}^i\boldsymbol{\Delta}+\boldsymbol{\Delta}^T{\boldsymbol{A}^i}^T\boldsymbol{A}^i\boldsymbol{\Delta}-2y_i\boldsymbol{A}^i\boldsymbol{\Delta}-(\boldsymbol{A}^i\boldsymbol{x^*})^{T}(\boldsymbol{A}^i\boldsymbol{x^*})\}}} \\
\end{multline}

\begin{equation} \label{eq:4}
    ={\dfrac{1}{\sigma^2}}{\sum^{m_3}_{i=m_2+1}{\{\dfrac{\boldsymbol{\Delta}^T{\boldsymbol{A}^i}^T\boldsymbol{A}^i\boldsymbol{\Delta}}{2}+\boldsymbol{x^*}^T{\boldsymbol{A}^i}^T\boldsymbol{A}^i\boldsymbol{\Delta}-y_i\boldsymbol{A}^i\boldsymbol{\Delta}\}}} \\
\end{equation}
\\
Again, define \\
$\mathcal{B}(\boldsymbol{x^*},\boldsymbol{\Delta})=<\boldsymbol{\Delta},\nabla L(\boldsymbol{x}^*)>$ \\
\begin{multline}
=\{-\dfrac{1}{\sigma}{\sum^{m_1}_{i=1}{\dfrac{\boldsymbol{A}^i\phi(\dfrac{\tau-\boldsymbol{A}^i\boldsymbol{x^*}}{\sigma})}{[1-\Phi(\dfrac{\tau-\boldsymbol{A}^i\boldsymbol{x^*}}{\sigma})]}}}
+\dfrac{1}{\sigma}{\sum^{m_2}_{i=m_1+1}{\dfrac{\boldsymbol{A}^i\phi(\dfrac{-\tau-\boldsymbol{A}^i\boldsymbol{x^*}}{\sigma})}{[\Phi(\dfrac{-\tau-\boldsymbol{A}^i\boldsymbol{x^*}}{\sigma})]}}}-\dfrac{1}{\sigma}{\sum^{m_3}_{i=m_2+1}{\boldsymbol{A}^i(\dfrac{y_i-\boldsymbol{A}^i\boldsymbol{x^*}}{\sigma})}}\}.\boldsymbol{\Delta}
\end{multline}
\begin{equation} \label{eq:5}
    =-\dfrac{1}{\sigma}{\sum^{m_1}_{i=1}{\dfrac{(\boldsymbol{A}^i\boldsymbol{\Delta})\phi(\dfrac{\tau-\boldsymbol{A}^i\boldsymbol{x^*}}{\sigma})}{[1-\Phi(\dfrac{\tau-\boldsymbol{A}^i\boldsymbol{x^*}}{\sigma})]}}}
+\dfrac{1}{\sigma}{\sum^{m_2}_{i=m_1+1}{\dfrac{(\boldsymbol{A}^i\boldsymbol{\Delta})\phi(\dfrac{-\tau-\boldsymbol{A}^i\boldsymbol{x^*}}{\sigma})}{[\Phi(\dfrac{-\tau-\boldsymbol{A}^i\boldsymbol{x^*}}{\sigma})]}}}-\dfrac{1}{\sigma^2}{\sum^{m_3}_{i=m_2+1}{({y_i\boldsymbol{A}^i\boldsymbol{\Delta}-\boldsymbol{x^*}^T{\boldsymbol{A}^i}^T\boldsymbol{A}^i\boldsymbol{\Delta}})}}
\end{equation}
\\ Now, $\delta L(\boldsymbol{x}^*,\boldsymbol{\Delta}) = \mathcal{A}(\boldsymbol{x^*},\boldsymbol{\Delta}) - \mathcal{B}(\boldsymbol{x^*},\boldsymbol{\Delta})$ \\
\begin{align*}
=-\sum^{m_1}_{i=1}({\ln{[1-\Phi(\frac{\tau-\boldsymbol{A}^{i}(\boldsymbol{x^*}+\boldsymbol{\Delta})}{\sigma})]}}-{\ln{[1-\Phi(\frac{\tau-\boldsymbol{A}^{i}\boldsymbol{x^*}}{\sigma})]}}) \\ -\sum^{m_2}_{i=m_1+1}({\ln{[\Phi(\frac{-\tau-\boldsymbol{A}^i(\boldsymbol{x^*}+\boldsymbol{\Delta})}{\sigma})]}}-{\ln{[\Phi(\frac{-\tau-\boldsymbol{A}^i\boldsymbol{x^*}}{\sigma})]}})\\
+{\dfrac{1}{\sigma^2}}{\sum^{m_3}_{i=m_2+1}{\{\dfrac{\boldsymbol{\Delta}^T{\boldsymbol{A}^i}^T\boldsymbol{A}^i\boldsymbol{\Delta}}{2}+\boldsymbol{x^*}^T{\boldsymbol{A}^i}^T\boldsymbol{A}^i\boldsymbol{\Delta}-y_i\boldsymbol{A}^i\boldsymbol{\Delta}\}}}
    +\dfrac{1}{\sigma}{\sum^{m_1}_{i=1}{\dfrac{(\boldsymbol{A}^i\boldsymbol{\Delta})\phi(\dfrac{\tau-\boldsymbol{A}^i\boldsymbol{x^*}}{\sigma})}{[1-\Phi(\dfrac{\tau-\boldsymbol{A}^i\boldsymbol{x^*}}{\sigma})]}}}\\
-\dfrac{1}{\sigma}{\sum^{m_2}_{i=m_1+1}{\dfrac{(\boldsymbol{A}^i\boldsymbol{\Delta})\phi(\dfrac{-\tau-\boldsymbol{A}^i\boldsymbol{x^*}}{\sigma})}{[\Phi(\dfrac{-\tau-\boldsymbol{A}^i\boldsymbol{x^*}}{\sigma})]}}}+\dfrac{1}{\sigma^2}{\sum^{m_3}_{i=m_2+1}{({y_i\boldsymbol{A}^i\boldsymbol{\Delta}-\boldsymbol{x^*}^T{\boldsymbol{A}^i}^T\boldsymbol{A}^i\boldsymbol{\Delta}})}} \\
\end{align*}
\begin{align} \label{eq:6}
  =\sum^{m_1}_{i=1}{\{-\ln{[1-\Phi(\frac{\tau-\boldsymbol{A}^{i}(\boldsymbol{x^*}+\boldsymbol{\Delta})}{\sigma})]}+\ln{[1-\Phi(\frac{\tau-\boldsymbol{A}^{i}\boldsymbol{x^*}}{\sigma})]}+\dfrac{(\boldsymbol{A^i}\boldsymbol{\Delta})\phi(\dfrac{\tau-\boldsymbol{A}^i\boldsymbol{x^*}}{\sigma})}{[1-\Phi(\dfrac{\tau-\boldsymbol{A}^i\boldsymbol{x^*}}{\sigma})]}\}}\nonumber \\
  +\sum^{m_2}_{i=m_1+1}{\{-\ln{[\Phi(\frac{-\tau-\boldsymbol{A}^i(\boldsymbol{x^*}+\boldsymbol{\Delta})}{\sigma})]}+\ln{[\Phi(\frac{-\tau-\boldsymbol{A}^i\boldsymbol{x^*}}{\sigma})]}-\dfrac{(\boldsymbol{A}^i\boldsymbol{\Delta})\phi(\dfrac{-\tau-\boldsymbol{A}^i\boldsymbol{x^*}}{\sigma})}{[\Phi(\dfrac{-\tau-\boldsymbol{A}^i\boldsymbol{x^*}}{\sigma})]}\}}\nonumber \\
  +\dfrac{1}{2\sigma^2}{\sum^{m_3}_{i=m_2+1}{\{\boldsymbol{\Delta}^T{\boldsymbol{A}^i}^T\boldsymbol{A}^i\boldsymbol{\Delta}\}}} \nonumber \\
\end{align}
$\delta L(\boldsymbol{x}^*,\boldsymbol{\Delta})$ consists of 3 terms as seen in equation  \ref{eq:6} :-\\
\begin{flalign} \label{eq:7}
    &\underline{\textbf{Term-1}} :\sum^{m_1}_{i=1}{\{-\ln{[1-\Phi(\frac{\tau-\boldsymbol{A}^{i}(\boldsymbol{x^*}+\boldsymbol{\Delta})}{\sigma})]}+\ln{[1-\Phi(\frac{\tau-\boldsymbol{A}^{i}\boldsymbol{x^*}}{\sigma})]}+{\dfrac{1}{\sigma}}\dfrac{(\boldsymbol{A}^i\boldsymbol{\Delta})\phi(\dfrac{\tau-\boldsymbol{A}^i\boldsymbol{x^*}}{\sigma})}{[1-\Phi(\dfrac{\tau-\boldsymbol{A}^i\boldsymbol{x^*}}{\sigma})]}\}}&\nonumber \\
    &\underline{\textbf{Term-2}} :\sum^{m_2}_{i=m_1+1}{\{-\ln{[\Phi(\frac{-\tau-\boldsymbol{A}^i(\boldsymbol{x^*}+\boldsymbol{\Delta})}{\sigma})]}+\ln{[\Phi(\frac{-\tau-\boldsymbol{A}^i\boldsymbol{x^*}}{\sigma})]}-{\dfrac{1}{\sigma}}\dfrac{(\boldsymbol{A}^i\boldsymbol{\Delta})\phi(\dfrac{-\tau-\boldsymbol{A}^i\boldsymbol{x^*}}{\sigma})}{[\Phi(\dfrac{-\tau-\boldsymbol{A}^i\boldsymbol{x^*}}{\sigma})]}\}}\nonumber &\\
    &\underline{\textbf{Term-3}} :\dfrac{1}{2\sigma^2}{\sum^{m_3}_{i=m_2+1}{\{\boldsymbol{\Delta}^T{\boldsymbol{A}^i}^T\boldsymbol{A}^i\boldsymbol{\Delta}\}}} \nonumber &\\
\end{flalign}
The target now is to prove that the  of each of the 3 terms separately and thereby prove the non-negativity of $\delta L(\boldsymbol{x^*},\boldsymbol{\Delta})$ . To do this, we try and prove the non-negativity of each term in equation \ref{eq:7} separately .\\
\subsubsection{To prove that $\textrm{Term1} \geq 0$}
Let us define a function:- \\
$g:\mathbb{R} \rightarrow \mathbb{R}$ such that $g(u)=\dfrac{\phi(u)}{1-\Phi(u)}$. \\
We rewrite \underline{\textbf{Term-1}} as follows:-\\
\begin{equation*}
\textrm{Term1} = \sum^{m_1}_{i=1}{\{-\ln{[1-\Phi(\frac{\tau-\boldsymbol{A}^{i}\boldsymbol{x^*}-\boldsymbol{A}^{i}\boldsymbol{\Delta})}{\sigma})]}+\ln{[1-\Phi(\frac{\tau-\boldsymbol{A}^{i}\boldsymbol{x^*}}{\sigma})]}+{\dfrac{1}{\sigma}}\dfrac{(\boldsymbol{A}^i\boldsymbol{\Delta})\phi(\dfrac{\tau-\boldsymbol{A}^i\boldsymbol{x^*}}{\sigma})}{[1-\Phi(\dfrac{\tau-\boldsymbol{A}^i\boldsymbol{x^*}}{\sigma})]}\}}.
\end{equation*}
Taking $u_i= \dfrac{\tau-\boldsymbol{A}^i\boldsymbol{x^*}}{\sigma}$ and $k_i=\dfrac{\boldsymbol{A}_i\boldsymbol{\Delta}}{\sigma}$  ,    $\forall i=1(1)m_1$, we can write Term-1 as:\\
\begin{equation} \label{eq:8}
    \sum^{m_1}_{i=1}{\{\ln{[1-\Phi(u_i)]}-\ln{[1-\Phi(u_i-k_i)]}+\dfrac{k_i\phi(u_i)}{[1-\Phi(u_i)]}\}} \\
\end{equation}
\\
\textbf{Defining a function :}\\
$f_1:\mathbb{R} \rightarrow \mathbb{R}$ such that , $f_1(u)=\ln{[1-\Phi(u)]}-\ln{[1-\Phi(u-k)]+kg(u)}$ ,where $k$ is any constant.\\ \\
\underline{\textbf{Claim-1 : $f_1(.)$ is a monotonically increasing function}} \\ \\
\underline{\textbf{Proof:}}
Differentiating $f_1$ w.r.t $u$, we get: \\
$f_1^{'}(u)=-\frac{\phi(u)}{1-\Phi(u)}+\frac{\phi(u-k)}{1-\Phi(u-k)}+k.g^{'}(u)$ 
\begin{flalign} \label{eq:9}
    & =-g(u)+g(u-k)+k.g^{'}(u) &\nonumber \\
\end{flalign}
Taking the Taylor's series expansion of g(.) up to the second term,\\
$g(u-k)=g(u)-\frac{k}{1!}g^{'}(u)+\frac{k^2}{2!}g''(\zeta)$ ; $\zeta \in (u-k,u) $ and $k \in \mathbb{R}$ \\
$\implies -g(u)+g(u-k)+k.g^{'}(u)=\frac{k^2}{2}g''(\zeta)$ \\
Replacing this form in the structure presented in equation \ref{eq:8}: \\
\begin{equation} \label{eq:10}
    f_1^{'}(u)=\frac{k^2}{2}g''(\zeta)
\end{equation}
\\
Here, $g(u)$ is the inverse of the Mills' ratio, which is proved to be a convex function in \cite{Gasull2014}. By definition of a strictly convex function, \\ 
\begin{equation} \label{eq:11}
g''(u) \geq 0 \quad \forall \quad u \in \mathbb{R}
\end{equation}
\\
Incorporating equation \ref{eq:10} in \ref{eq:9} , we get, for any $k=k_i \quad \forall i=1,2,...m$ \\ $f_1^{'}(u)=\frac{k^2}{2}g''(\zeta) > \dfrac{ \|\boldsymbol{A}^i\boldsymbol{\Delta}\|_2^2}{\sigma^2}.g''(\zeta) \geq 0 \quad \forall u \in \rm I \!R$ .\\
Hence, $f_1^{'}(u) \geq 0 \quad \forall u \in \rm I \!R $. \\ This implies that $\boldsymbol{f_1(.)}$ \textbf{\underline{is a monotonically increasing function}}.
\\ \\
\underline{\textbf{Claim-2 : $f_1(-\infty) = 0$ }} \\ \\
\underline{\textbf{Proof:}}
Now,
$\lim_{u \to -\infty} \Phi(u-k) = \lim_{u \to -\infty} \Phi(u) = 0$ \\
$\implies \ln[1-\Phi(-\infty)] = 0 \\ \implies \lim_{u \to -\infty}\ln[1-\Phi(u-k)] = \lim_{u \to -\infty}\ln[1-\Phi(u)] = 0$ \\
Also, $\lim_{u \to -\infty}\phi(u) = 0 \quad \implies \lim_{u \to -\infty}g(u)=0$
\\ \\
\begin{multline}
    f_1(-\infty)= \lim_{u \to -\infty} f_1(u) 
    = \lim_{u \to -\infty} [\ln{[1-\Phi(u)]}-\ln{[1-\Phi(u-k)]+kg(u)}] \\
    = \lim_{u \to -\infty} \ln[1-\Phi(u)] - \lim_{u \to -\infty} \ln[1-\Phi(u-k)] + k \lim_{u \to -\infty} g(u)
    = 0
\end{multline}
\\
Hence, $\boldsymbol{\underline{f_1(-\infty)=0}}$

Thus, from Claim-1 and Claim-2 , $f_1(.)$ is a monotonically increasing function bounded below by 0 . This implies, 
\begin{equation} \label{eq:12}
    f_1(u) \geq 0 \quad \forall \quad u \in \mathbb{R}
\end{equation} \\
Putting equation \ref{eq:12} in \ref{eq:8}, we have, \\ \\
Term 1 $= \sum^{m_1}_{i=1}{f_1(u_i)} \quad $. Since $f_1(u_i) \geq 0 \quad \forall \quad u_i$ ,\\
\begin{equation} \label{eq:13}
    \textbf{Term 1} \geq 0
\end{equation}

\subsubsection{To prove that $\textrm{Term2} \geq 0$}
Let us define a function:- \\
$h:\mathbb{R} \rightarrow \mathbb{R}$ such that $h(v)=\dfrac{\phi(v)}{\Phi(v)}$. \\
Rewriting Term 2 as follows: \\
\textbf{Term 2}=
\begin{align*}
    \sum^{m_2}_{i=m_1+1}{\{-\ln{[\Phi(\frac{-\tau-\boldsymbol{A}^i\boldsymbol{x^*}-\boldsymbol{A}^i\boldsymbol{\Delta}}{\sigma})]}+\ln{[\Phi(\frac{-\tau-\boldsymbol{A}^i\boldsymbol{x^*}}{\sigma})]}-{\dfrac{1}{\sigma}}\dfrac{(\boldsymbol{A}^i\boldsymbol{\Delta})\phi(\dfrac{-\tau-\boldsymbol{A}^i\boldsymbol{x^*}}{\sigma})}{[\Phi(\dfrac{-\tau-\boldsymbol{A}^i\boldsymbol{x^*}}{\sigma})]}\}} \\
\end{align*}
Taking $v_i=\frac{-t-\boldsymbol{A}^i\boldsymbol{x^*}}{\sigma} \quad and \quad k_i=\frac{\boldsymbol{A}^i\boldsymbol{\Delta}}{\sigma}$, we have Term 2 as: 
\begin{equation} \label{eq:14}
    \sum^{m_2}_{i=m_1+1}{\{\ln[\Phi(v_i)]-\ln[\Phi(v_i-k_i)]-k_i.\frac{\phi(v_i)}{\Phi(v_i)}\}}
\end{equation}
\\
\textbf{Defining a function :}\\
$f_2:\mathbb{R} \rightarrow \mathbb{R}$ such that , $f_2(v)=\ln{[\Phi(v)]}-\ln{[\Phi(v-k)]+k.h(v)}$ ,where k is any constant.\\ \\
\underline{\textbf{Claim-3 : $f_2(.)$ is a monotonically decreasing function}} \\ \\
\underline{\textbf{Proof:}}
Differentiating $f_2$ w.r.t $v$, we get: \\
$f_2^{'}(v)=\frac{\phi(v)}{\Phi(v)}-\frac{\phi(v-k)}{\Phi(v-k)}-k.h^{'}(v)$
\begin{flalign} \label{eq:15}
    &=h(v)-h(v-k)-k.h^{'}(v)& \nonumber\\
\end{flalign}
\\
Taking the Taylor's Series expansion of $h(.)$ up to the second term,\\
$h(v-k)=h(v)-\frac{k}{1!}h^{'}(v)+\frac{k^2}{2!}h^{''}(\zeta)$ ; $\zeta \in (v-k,v) $ and $k \in \mathbb{R}$ \\
$\implies h(v)-h(v-k)-k.h^{'}(v)=-\frac{k^2}{2}h^{''}(\zeta)$ \\
\begin{equation} \label{eq:16}
    f_2^{'}(v)=-\frac{k^2}{2}h^{''}(\zeta)
\end{equation}
\textbf{Lemma-1: h(.) is a convex function} \\
\textbf{Proof:} 
Related to standard normal, consider two  properties:- \\
\begin{multline}
 1) \phi(x)=\phi(-x) \quad \forall x \in \mathbb{R} \\ 
 2) 1-\Phi(x)=\Phi(-x) \quad \forall x \in \mathbb{R} \\
\end{multline}
We have,\\ $g(x) = \frac{\phi(x)}{{1-\Phi(x)}} = \frac{\phi(-x)}{\Phi(-x)} = h(-x) \quad \forall x \in \mathbb{R} \quad \implies g(x)=h(-x) \quad  \forall x \in \mathbb{R}$ \\
Differentiating w.r.t x,    
$\quad g^{'}(x)=-h^{'}(-x) \quad \forall x \in \mathbb{R}$ \\
Again, differentiating w.r.t x, $\quad g''(x)=-(-h^{''}(-x)) = h^{''}(-x) \quad \forall x \in \mathbb{R}$ \\ $\implies h^{''}(-x)=g''(x) \quad \forall x \in \mathbb{R}$ as g(.) is a convex function. \\
So, $h^{''}(-x) \geq 0 \implies h^{''}(x) \geq 0 \quad \forall x \in \mathbb{R}$\\
\begin{equation} \label{eq:17}
    h(.) \quad \textrm{is a convex function}
\end{equation}
Thus putting equation \ref{eq:16} in \ref{eq:15}, we get, for any $k=k_i \quad \forall i=1,2,...m$ \\ $f_2^{'}(v)=-\frac{k^2}{2}h^{''}(\zeta) \leq \dfrac{ \|\boldsymbol{A^i}\boldsymbol{\Delta}\|_2^2}{\sigma^2}.h^{''}(\zeta) \leq 0 \quad \forall v \in \rm I \!R$ .\\
Hence, $f_2^{'}(v) \leq 0 \quad \forall v \in \rm I \!R$.\\ This implies that $\boldsymbol{f_2(.)}$ \textbf{\underline{is a monotonically decreasing function}}.
\\ \\
\underline{\textbf{Claim-4 : $f_2(\infty) = 0$ }} \\ \\
\underline{\textbf{Proof:}}
Now,
$\lim_{v \to \infty} \Phi(v-k) = \lim_{v \to v\infty} \Phi(v) = 1$ \\
$\implies \ln[\Phi(\infty)] = 0 \\ \implies \lim_{v \to \infty}\ln[\Phi(v-k)] = \lim_{v \to \infty}\ln[\Phi(v)] = 0$ \\
Also, $\lim_{v \to \infty}\phi(v) = 0 \quad \implies \lim_{v \to \infty}h(v)=0$
\\ \\
\begin{multline}
    f_2(\infty)= \lim_{v \to v\infty} f_2(v) 
    = \lim_{v \to \infty} [\ln{[\Phi(v)]}-\ln{[\Phi(v-k)]-k.h(v)}] \\
    = \lim_{v \to \infty} \ln[\Phi(v)] - \lim_{v \to \infty} \ln[\Phi(v-k)] - k \lim_{v \to v\infty} h(v)
    = 0
\end{multline}
\\
Hence, $\boldsymbol{\underline{f_2(\infty)=0}}$
\\
Thus, from Claim-1 and Claim-2 , $f_2(.)$ is a monotonically decreasing function bounded below by 0 . This implies, 
\begin{equation} \label{eq:18}
    f_2(v) \geq 0 \quad \forall \quad v \in \mathbb{R}
\end{equation} \\
Putting equation \ref{eq:18} in equation \ref{eq:14}, we have, \\ \\
Term 2 $= \sum^{m_2}_{i=m_1+1}{f_2(v_i)} \quad $. Since $f_2(v_i) \geq 0 \quad \forall \quad v_i$ ,\\
\begin{equation} \label{eq:19}
    \textbf{Term 2} \geq 0
\end{equation}
\subsubsection{To prove that $\textrm{Term3} \geq 0$}
We can write the matrix $\boldsymbol{A}$ as,\\
$\boldsymbol{A}^{m \cross n}$=
$
\begin{bmatrix}
\boldsymbol{A_1}^{m_1 \cross n} \\
\boldsymbol{A_2}^{m_2 \cross n} \\
\boldsymbol{A_3}^{m_3 \cross n} \\
\end{bmatrix}
$
\\
Rewriting Term 3 as follows:\\
$\boldsymbol{Term 3}= \frac{1}{2\sigma^2}{\sum^{m_3}_{i=m_2+1}{\{\boldsymbol{\Delta}^T{\boldsymbol{A}^i}^T\boldsymbol{A}^i\boldsymbol{\Delta}\}}}=\frac{1}{2\sigma^2}{\{\boldsymbol{\Delta}^T\boldsymbol{A}_3^T\boldsymbol{A}_3\boldsymbol{\Delta}\}}$ \\
\\
Also, $\sum^{m_3}_{i=m_2+1}{k_i^2}=\sum^{m_3}_{i=m_2+1}{\frac{(\boldsymbol{A}^i\boldsymbol{\Delta})^T(\boldsymbol{A}^i\boldsymbol{\Delta})}{\sigma^2}} =\dfrac{ \boldsymbol{\Delta}^T\boldsymbol{A}_3^T\boldsymbol{A}_3\boldsymbol{\Delta}}{\sigma^2} $ \\ \\
From \cite{THW2015}, $\boldsymbol{\Delta}^T\boldsymbol{A}_3^T\boldsymbol{A}_3\boldsymbol{\Delta}=\|\boldsymbol{A}_3\boldsymbol{\Delta}\|_2^2 > \gamma \|\boldsymbol{\Delta}\|_2^2 \quad \forall \boldsymbol{\Delta} \in \textbf{C} \quad ; \quad \\where, \quad \textbf{C}=\{\boldsymbol{\Delta} \in \mathbb{R} |\quad \|\boldsymbol{\Delta}_{\overline{S}}\|_1 \leq \alpha\|\boldsymbol{\Delta}_S\|_1\}$ and $\gamma$ is non-negative constant. \\ \\
So,
\begin{equation} \label{eq:20}
    \sum^{m_3}_{i=m_2+1}{k_i^2} > \dfrac{\gamma \|\boldsymbol{\Delta}\|_2^2}{\sigma^2} 
\end{equation}
From \ref{eq:20}, we have,\\
$\textbf{Term 3}=\dfrac{1}{2}{ \sum^{m_3}_{i=m_2+1}{k_i^2}}>\dfrac{\gamma \|\boldsymbol{\Delta}\|_2^2}{2\sigma^2} \geq 0$ \\
Thus,
\begin{equation}\label{eq:21}
    \textbf{Term 3} > 0
\end{equation}
\subsubsection{$L(\boldsymbol{y},\boldsymbol{Ax};\tau)$ satisfies the RSC property}
Thus, from equations \ref{eq:13}, \ref{eq:19} and \ref{eq:21}, we have, \\
$\delta L(\boldsymbol{x^*},\boldsymbol{\Delta})= \textrm{Term 1} + \textrm{Term 2} + \textrm{Term 3} \geq \dfrac{\gamma \|\boldsymbol{\Delta}\|_2^2}{2\sigma^2} $ \\
This inequality holds for $\boldsymbol{\Delta} \in C$ where $C \triangleq \{ \boldsymbol{\Delta} |\quad \|\boldsymbol{\Delta}_{S^c}\|_1 \leq \alpha\|\boldsymbol{\Delta}_S\|_1 \}$ \\
In our model, the vector $\boldsymbol{x^*}$ is strictly sparse. Hence, $\|\boldsymbol{x}_{true_{S^c}}\|_1=0$. \\
Taking $\alpha = 3$ , the set $C$ satisfies the condition on $\boldsymbol{\Delta}$ require by RSC.\\
Hence, L(x) satisfies Restricted Strong Convexity with curvature $\kappa_L= \dfrac{\gamma}{2\sigma^2}$.

\subsection{Theorem 3: Lower Bound on the gradient of the loss function}
\subsubsection{The Gradient of the Cost Function}
The gradient term represented by $\nabla L$ is shown by:
\begin{equation} \label{eq:22}
    \nabla L = -\dfrac{1}{\sigma}{\sum^{m_1}_{i=1}{\dfrac{\boldsymbol{A}^i\phi(\dfrac{\tau-\boldsymbol{A^i}\boldsymbol{x^*}}{\sigma})}{[1-\Phi(\dfrac{\tau-\boldsymbol{A}^i\boldsymbol{x^*}}{\sigma})]}}}
+\dfrac{1}{\sigma}{\sum^{m_2}_{i=m_1+1}{\dfrac{\boldsymbol{A}^i\phi(\dfrac{-\tau-\boldsymbol{A}^i\boldsymbol{x^*}}{\sigma})}{[\Phi(\dfrac{-\tau-\boldsymbol{A}^i\boldsymbol{x^*}}{\sigma})]}}}-\dfrac{1}{\sigma}{\sum^{m_3}_{i=m_2+1}{\boldsymbol{A}^i(\dfrac{y_i-\boldsymbol{A}^i\boldsymbol{x^*}}{\sigma})}}
\end{equation}
$\nabla L$ consists of 3 terms:-
\begin{eqnarray} \label{eq:23}
\textbf{Term 1:} = \frac{1}{\sigma}{\sum^{m_1}_{i=1}{\dfrac{\boldsymbol{A}^i\phi(\dfrac{\tau-\boldsymbol{A}^i\boldsymbol{x^*}}{\sigma})}{[1-\Phi(\dfrac{\tau-\boldsymbol{A}^i\boldsymbol{x^*}}{\sigma})]}}} \nonumber \\
\textbf{Term 2:} = \dfrac{1}{\sigma}{\sum^{m_2}_{i=m_1+1}{\dfrac{\boldsymbol{A}^i\phi(\dfrac{-\tau-\boldsymbol{A}^i\boldsymbol{x^*}}{\sigma})}{[\Phi(\dfrac{-\tau-\boldsymbol{A}^i\boldsymbol{x^*}}{\sigma})]}}} \nonumber \\
\textbf{Term 3:} =-\dfrac{1}{\sigma}{\sum^{m_3}_{i=m_2+1}{\boldsymbol{A}^i(\dfrac{y_i-\boldsymbol{A}^i\boldsymbol{x^*}}{\sigma})}}
\end{eqnarray}
such that $\nabla L = - \textrm{Term 1+Term 2 + Term 3}$ 
\subsubsection{A necessary condition}
For deriving this bound, we consider one condition :\\
The signal $\boldsymbol{x}$ is bounded i.e., 
\begin{equation} \label{eq:24}
    \boldsymbol{\alpha} \leq \boldsymbol{x} \leq \boldsymbol{\beta}    
\end{equation}
, where all elements of $\boldsymbol{\alpha}$ is $\alpha$ and all elements of $\boldsymbol{\beta}$ is $\beta$.
\subsubsection{Bounds on $\Phi(.)$ and $\phi(.)$}
From (3) , we have, $\alpha \leq x_j \leq \beta \quad \forall \, j=1,2,....,n$. Thus for any $i=1,2,....,m$, ($A^{ij}$ being the $<i,j>^{th}$ element of $\boldsymbol{A}$ )\\
If $A^{ij} \geq 0$, then $A^{ij}\alpha \leq A^{ij}x_j \leq A^{ij}\beta \quad \forall j:A^{ij} \geq 0$ \\
Again if $A^{ij} < 0$, then $A^{ij}\beta \leq A^{ij}x_j \leq A^{ij}\alpha \quad \forall j:A^{ij} < 0$.\\
We have, $ \boldsymbol{A}^i\boldsymbol{x} = \sum^{n}_{j=1}{A^{ij}}x_j \quad \forall i=1,2,....,m$. So, $\boldsymbol{A}^i\boldsymbol{x}$ is bounded by,
\begin{equation}\label{eq:25}
    \sum_{j:A^{ij} \geq 0}{A^{ij}\alpha} + \sum_{j:A^{ij} < 0}{A^{ij}\beta} \leq \sum^{n}_{j=1}{A^{ij}x_j} \leq \sum_{j:A^{ij} \geq 0}{A^{ij}\beta} + \sum_{j:A^{ij} < 0}{A^{ij}\alpha}
\end{equation}
Let $p_i = \sum_{j:A^{ij} \geq 0}{A^{ij}\alpha} + \sum_{j:A^{ij} < 0}{A^{ij}\beta} \quad and\quad  q_i=\sum_{j:A^{ij} \geq 0}{A^{ij}\beta} + \sum_{j:A^{ij} < 0}{A^{ij}\alpha} \quad \forall i=1,2,...,m$. Thus we have,
\begin{eqnarray}\label{eq:26}
    p_i \leq \boldsymbol{A}^i\boldsymbol{x} \leq q_i \nonumber \\ \implies \dfrac{\tau-q_i}{\sigma} \leq \dfrac{\tau-\boldsymbol{A}^i\boldsymbol{x}}{\sigma} \leq \dfrac{\tau-p_i}{\sigma} 
\end{eqnarray}
We know that ,$\Phi(.)$ is a non-decreasing function. Thus, from (5) , $\forall i=1,2,....,m$,
\begin{eqnarray}\label{eq:27}
    \Phi(\dfrac{\tau-q_i}{\sigma}) \leq \Phi(\dfrac{\tau-\boldsymbol{A}^i\boldsymbol{x}}{\sigma}) \leq \Phi(\dfrac{\tau-p_i}{\sigma}) \nonumber \\ \implies \dfrac{1}{\Phi(\dfrac{\tau-p_i}{\sigma})} \leq \dfrac{1}{\Phi(\dfrac{\tau-\boldsymbol{A}^i\boldsymbol{x}}{\sigma})} 
    \leq \dfrac{1}{\Phi(\dfrac{\tau-q_i}{\sigma})} \\
    Also, \quad 1-\Phi(\dfrac{\tau-p_i}{\sigma}) \leq \label{eq:28} 1-\Phi(\dfrac{\tau-\boldsymbol{A}^i\boldsymbol{x}}{\sigma}) \leq 1-\Phi(\dfrac{\tau-q_i}{\sigma}) \nonumber \\
    \implies \dfrac{1}{1-\Phi(\dfrac{\tau-q_i}{\sigma})} \leq \dfrac{1}{1-\Phi(\dfrac{\tau-\boldsymbol{A}^i\boldsymbol{x}}{\sigma})} 
    \leq \dfrac{1}{1-\Phi(\dfrac{\tau-p_i}{\sigma})} 
\end{eqnarray}
Now, since $\phi(.)$ is not a monotone function, there are 3 cases depending on the values of $p_i$ and $q_i$ , $\forall i=1,2,....,m$.\\
\textbf{Case 1:} $-\infty < \tau-q_i \leq \tau-p_i \leq 0 \quad \textrm{for some i}$ \\
$\phi(.)$ is an increasing function in $(-\infty,0]$ . Hence, 
\begin{eqnarray} \label{eq:29}
    \phi(\dfrac{\tau-q_i}{\sigma}) \leq \phi(\dfrac{\tau-\boldsymbol{A}^i\boldsymbol{x}}{\sigma}) \leq \phi(\dfrac{\tau-p_i}{\sigma}) \nonumber \\ \implies K_i \leq \phi(\dfrac{\tau-\boldsymbol{A}^i\boldsymbol{x}}{\sigma}) \leq L_i \quad \forall i:-\infty < \tau-q_i \leq \tau-p_i \leq 0
\end{eqnarray}
where, $K_i=\phi(\dfrac{\tau-q_i}{\sigma})$ and $L_i=\phi(\dfrac{\tau-p_i}{\sigma})$.\\ 
\textbf{Case 2:} $0 \leq \tau-q_i \leq \tau-p_i < \infty \quad \textrm{for some i}$ \\
$\phi(.)$ is a non-increasing function on $[0,\infty)$. Hence,
\begin{eqnarray}\label{eq:30}
    \phi(\dfrac{\tau-p_i}{\sigma}) \leq \phi(\dfrac{\tau-\boldsymbol{A}^i\boldsymbol{x}}{\sigma}) \leq \phi(\dfrac{\tau-q_i}{\sigma}) \nonumber \\ \implies K_i \leq \phi(\dfrac{\tau-\boldsymbol{A}^i\boldsymbol{x}}{\sigma}) \leq L_i \quad \forall i:0 \leq \tau-q_i \leq \tau-p_i < \infty
\end{eqnarray}
where, $K_i=\phi(\dfrac{\tau-p_i}{\sigma})$ and $L_i=\phi(\dfrac{\tau-q_i}{\sigma})$.\\
\textbf{Case 3:} $-\infty < \tau-q_i \leq 0 \leq \tau-p_i < \infty \quad \textrm{for some i}$\\
Here,
\begin{eqnarray} \label{eq:31}
    min\{\phi(\dfrac{\tau-q_i}{\sigma}),\phi(\dfrac{\tau-p_i}{\sigma})\} \leq \phi(\dfrac{\tau-\boldsymbol{A}^i\boldsymbol{x}}{\sigma}) \leq \phi(0) =\dfrac{1}{\sqrt{2\pi}} \nonumber \\
    \implies K_i \leq \phi(\dfrac{\tau-\boldsymbol{A}^i\boldsymbol{x}}{\sigma}) \leq L_i \quad \forall i:-\infty < \tau-q_i \leq 0 \leq \tau-p_i < \infty
\end{eqnarray}
where $K_i=min\{\phi(\dfrac{\tau-q_i}{\sigma}),\phi(\dfrac{\tau-p_i}{\sigma})\}$ and $L_i=\dfrac{1}{\sqrt{2\pi}}$.
\subsubsection{Bound for Term 1}
Combining equations \ref{eq:28} ,\ref{eq:29}, \ref{eq:30} and \ref{eq:31} together, we get,
\begin{equation}\label{eq:32}
    \dfrac{K_i}{1-\Phi(\dfrac{\tau-q_i}{\sigma})} \leq \dfrac{\phi(\dfrac{\tau-\boldsymbol{A}^i\boldsymbol{x}}{\sigma})}{1-\Phi(\dfrac{\tau-\boldsymbol{A}^i\boldsymbol{x}}{\sigma})} 
    \leq \dfrac{L_i}{1-\Phi(\dfrac{\tau-p_i}{\sigma})} \quad \forall \, i=1,2,...,m_1
\end{equation}
For a given $\boldsymbol{A^i}$, if ${A^{ij}}$ ,i.e. the $j^{th}$ element of the row vector $\boldsymbol{A^i}$ of the sensing matrix $\boldsymbol{A}$ is positive, then multiplying ${A^{ij}}$ to equation \ref{eq:32} , we get ,
\begin{eqnarray} \label{eq:33}
    \dfrac{A^{ij}K_i}{1-\Phi(\dfrac{\tau-q_i}{\sigma})} \leq \dfrac{A^{ij}\phi(\dfrac{\tau-\boldsymbol{A}^i\boldsymbol{x}}{\sigma})}{1-\Phi(\dfrac{\tau-\boldsymbol{A}^i\boldsymbol{x}}{\sigma})} 
    \leq \dfrac{A^{ij}L_i}{1-\Phi(\dfrac{\tau-p_i}{\sigma})} \quad \forall \, i=1,2,...,m_1 \nonumber \\
    \implies 
    U^{ij} \leq \dfrac{A^{ij}\phi(\dfrac{\tau-\boldsymbol{A}^i\boldsymbol{x}}{\sigma})}{1-\Phi(\dfrac{\tau-\boldsymbol{A}^i\boldsymbol{x}}{\sigma})} \leq V^{ij} \quad \forall \, i=1,2,...,m_1
\end{eqnarray}
where $U^{ij}=\dfrac{A^{ij}K_i}{1-\Phi(\dfrac{\tau-q_i}{\sigma})}$ and $V^{ij}=\dfrac{A^{ij}L_i}{1-\Phi(\dfrac{\tau-p_i}{\sigma})}$ \\
Again if, ${A^{ij}}$ is negative, then multiplying ${A^{ij}}$ to equation \ref{eq:32} , we get ,
\begin{eqnarray} \label{eq:34}
    \dfrac{A^{ij}L_i}{1-\Phi(\dfrac{\tau-p_i}{\sigma})} \leq \dfrac{A^{ij}\phi(\dfrac{\tau-\boldsymbol{A}^i\boldsymbol{x}}{\sigma})}{1-\Phi(\dfrac{\tau-\boldsymbol{A}^i\boldsymbol{x}}{\sigma})} 
    \leq \dfrac{A^{ij}K_i}{1-\Phi(\dfrac{\tau-q_i}{\sigma})} \quad \forall \, i=1,2,...,m_1 \nonumber \\
    \implies 
    U^{ij} \leq \dfrac{A^{ij}\phi(\dfrac{\tau-\boldsymbol{A}^i\boldsymbol{x}}{\sigma})}{1-\Phi(\dfrac{\tau-\boldsymbol{A}^i\boldsymbol{x}}{\sigma})} \leq V^{ij} \quad \forall \, i=1,2,...,m_1
\end{eqnarray}
where $U^{ij}=\dfrac{A^{ij}L_i}{1-\Phi(\dfrac{\tau-p_i}{\sigma})}$ and $V^{ij}=\dfrac{A^{ij}K_i}{1-\Phi(\dfrac{\tau-q_i}{\sigma})}$ \\
Let us define $\boldsymbol{U}^i=(U^{i1},U^{i2},.....,U^{in})$ and $\boldsymbol{V}^i=(V^{i1},V^{i2},.....,V^{in})$ .We can now join equations \ref{eq:33} and \ref{eq:34} as a vector inequality as follows,
\begin{eqnarray} \label{eq:35}
    \boldsymbol{U}^i \leq \dfrac{\boldsymbol{A}^{i}\phi(\dfrac{\tau-\boldsymbol{A}^i\boldsymbol{x}}{\sigma})}{1-\Phi(\dfrac{\tau-\boldsymbol{A}^i\boldsymbol{x}}{\sigma})} \leq \boldsymbol{V}^i \quad \forall \, i=1,2,...,m_1
\end{eqnarray}
From equation \ref{eq:35}, summing over $i=1,2,...,m_1$ and multiplying throughout by $\dfrac{1}{\sigma}$, we have, 
\begin{equation} \label{eq:36}
    \dfrac{1}{\sigma}\sum^{m_1}_{i=1}{\boldsymbol{U}^i} \leq \dfrac{1}{\sigma}\sum^{m_1}_{i=1}{\dfrac{\boldsymbol{A}^{i}\phi(\dfrac{\tau-\boldsymbol{A}^i\boldsymbol{x}}{\sigma})}{1-\Phi(\dfrac{\tau-\boldsymbol{A}^i\boldsymbol{x}}{\sigma})}} \leq \dfrac{1}{\sigma}\sum^{m_1}_{i=1}{\boldsymbol{V}^i}
\end{equation}
These are the bounds for \textbf{Term 1}
\subsubsection{Bound for Term 2}
In equations \ref{eq:27}. \ref{eq:29},\ref{eq:30} and \ref{eq:31},  replace $\tau$ by $-\tau$. Let $\overline{K_i}= K_i$ with $\tau$ replaced by $-\tau$ and $\overline{L_i}= L_i$ with $\tau$ replaced by $-\tau$ $\forall i=1,2,...m_2$. Now combining these equations together, we have,
\begin{equation} \label{eq:37}
    \dfrac{\overline{K_i}}{1-\Phi(\dfrac{-\tau-p_i}{\sigma})} \leq \dfrac{\phi(\dfrac{-\tau-\boldsymbol{A}^i\boldsymbol{x}}{\sigma})}{\Phi(\dfrac{-\tau-\boldsymbol{A}^i\boldsymbol{x}}{\sigma})} 
    \leq \dfrac{\overline{L_i}}{\Phi(\dfrac{-\tau-q_i}{\sigma})} \quad \forall \, i=1,2,...,m_2
\end{equation}
For a given $\boldsymbol{A^i}$, if ${A^{ij}}$ ,i.e. the $j^{th}$ element of the vector $\boldsymbol{A^i}$, is positive, then multiplying ${A^{ij}}$ to \ref{eq:37} , we get ,
\begin{eqnarray} \label{eq:38}
    \dfrac{A^{ij}\overline{K_i}}{1-\Phi(\dfrac{-\tau-q_i}{\sigma})} \leq \dfrac{A^{ij}\phi(\dfrac{-\tau-\boldsymbol{A}^i\boldsymbol{x}}{\sigma})}{1-\Phi(\dfrac{-\tau-\boldsymbol{A}^i\boldsymbol{x}}{\sigma})} 
    \leq \dfrac{A^{ij}\overline{L_i}}{1-\Phi(\dfrac{-p_i}{\sigma})} \quad \forall \, i=1,2,...,m_2 \nonumber \\
    \implies 
    \overline{U^{ij}} \leq \dfrac{A^{ij}\phi(\dfrac{-\tau-\boldsymbol{A}^i\boldsymbol{x}}{\sigma})}{1-\Phi(\dfrac{-\tau-\boldsymbol{A}^i\boldsymbol{x}}{\sigma})} \leq \overline{V^{ij}} \quad \forall \, i=1,2,...,m_2
\end{eqnarray}
where $\overline{U^{ij}}=\dfrac{A^{ij}\overline{K_i}}{1-\Phi(\dfrac{-\tau-q_i}{\sigma})}$ and $\overline{V^{ij}}=\dfrac{A^{ij}\overline{L_i}}{1-\Phi(\dfrac{-\tau-p_i}{\sigma})}$ \\
Again if, ${A^{ij}}$ is negative, then multiplying ${A^{ij}}$ to equation \ref{eq:37} , we get ,
\begin{eqnarray}\label{eq:39}
    \dfrac{A^{ij}\overline{L_i}}{1-\Phi(\dfrac{-\tau-p_i}{\sigma})} \leq \dfrac{A^{ij}\phi(\dfrac{-\tau-\boldsymbol{A}^i\boldsymbol{x}}{\sigma})}{1-\Phi(\dfrac{-\tau-\boldsymbol{A}^i\boldsymbol{x}}{\sigma})} 
    \leq \dfrac{A^{ij}\overline{K_i}}{1-\Phi(\dfrac{-\tau-q_i}{\sigma})} \quad \forall \, i=1,2,...,m_2 \nonumber \\
    \implies 
    \overline{U^{ij}} \leq \dfrac{A^{ij}\phi(\dfrac{-\tau-\boldsymbol{A}^i\boldsymbol{x}}{\sigma})}{1-\Phi(\dfrac{\tau-\boldsymbol{A}^i\boldsymbol{x}}{\sigma})} \leq \overline{V^{ij}} \quad \forall \, i=1,2,...,m_2
\end{eqnarray}
where $\overline{U^{ij}}=\dfrac{A^{ij}\overline{L_i}}{1-\Phi(\dfrac{-\tau-p_i}{\sigma})}$ and $\overline{V^{ij}}=\dfrac{A^{ij}\overline{K_i}}{1-\Phi(\dfrac{-\tau-q_i}{\sigma})}$ \\
Let us define $\overline{\boldsymbol{U}^i}=(\overline{U^{i1}},\overline{U^{i2}},.....,\overline{U^{in}})$ and $\overline{\boldsymbol{V}^i}=(\overline{V^{i1}},\overline{V^{i2}},.....,\overline{V^{in}})$ .We can now join equations \ref{eq:38} and \ref{eq:39} as a vector inequality as follows,
\begin{eqnarray} \label{eq:40}
    \overline{\boldsymbol{U}^i} \leq \dfrac{\boldsymbol{A}^{i}\phi(\dfrac{-\tau-\boldsymbol{A}^i\boldsymbol{x}}{\sigma})}{1-\Phi(\dfrac{-\tau-\boldsymbol{A}^i\boldsymbol{x}}{\sigma})} \leq \overline{\boldsymbol{V}^i} \quad \forall \, i=1,2,...,m_2
\end{eqnarray}
From equation \ref{eq:40}, summing over $i=m_1+1,2,...,m_2$ and multiplying throughout by $\dfrac{1}{\sigma}$, we have, 
\begin{equation} \label{eq:41}
    \dfrac{1}{\sigma}\sum^{m_2}_{i=m_1+1}{\overline{\boldsymbol{U}^i}} \leq \dfrac{1}{\sigma}\sum^{m_2}_{i=m_1+1}{\dfrac{\boldsymbol{A}^{i}\phi(\dfrac{-\tau-\boldsymbol{A}^i\boldsymbol{x}}{\sigma})}{1-\Phi(\dfrac{-\tau-\boldsymbol{A}^i\boldsymbol{x}}{\sigma})}} \leq \dfrac{1}{\sigma}\sum^{m_2}_{i=m_1+1}{\overline{\boldsymbol{V}^i}}
\end{equation}
These are the bounds for \textbf{Term 2}
\subsubsection{$L_{\infty}$ norm bounds for Term2-Term1}
From equation \ref{eq:36} and \ref{eq:41}, \textbf{Term 2 - Term 1} is bounded by,
\begin{eqnarray} \label{eq:42}
    \frac{1}{\sigma}\{\sum^{m_1}_{i=1}{\boldsymbol{V}^i}+\sum^{m_2}_{i=m_1+1}{\overline{\boldsymbol{U}^i}}\} \nonumber \\
    \leq -\dfrac{1}{\sigma}\sum^{m_1}_{i=1}{\dfrac{\boldsymbol{A}^{i}\phi(\dfrac{\tau-\boldsymbol{A}^i\boldsymbol{x}}{\sigma})}{1-\Phi(\dfrac{\tau-\boldsymbol{A}^i\boldsymbol{x}}{\sigma})}}+\dfrac{1}{\sigma}\sum^{m_2}_{m_1+1}{\dfrac{\boldsymbol{A}^{i}\phi(\dfrac{-\tau-\boldsymbol{A}^i\boldsymbol{x}}{\sigma})}{\Phi(\dfrac{-\tau-\boldsymbol{A}^i\boldsymbol{x}}{\sigma})}} \nonumber \\
    \leq \dfrac{1}{\sigma}\{\sum^{m_1}_{i=1}{\boldsymbol{U}^i}+\sum^{m_2}_{i=m_1+1}{\overline{\boldsymbol{V}^i}}\}
\end{eqnarray}
From the inequality \ref{eq:42}, the $L_{\infty}$ norm on \textbf{Term 2 - Term 1} would be bound by,
\begin{eqnarray} \label{eq:43}
    \|\textbf{Term 2 - Term 1}\|_{\infty} \leq \frac{1}{\sigma} \textrm{max}\{\sum^{m_1}_{i=1}{\boldsymbol{V}^i}+\sum^{m_2}_{i=m_1+1}{\overline{\boldsymbol{U}^i}},\sum^{m_1}_{i=1}{\boldsymbol{U}^i}+\sum^{m_2}_{i=m_1+1}{\overline{\boldsymbol{V}^i}}\} \nonumber \\ 
    \leq \dfrac{Q}{\sigma}
\end{eqnarray}
where we define $Q \triangleq \textrm{max}\{\sum^{m_1}_{i=1}{\boldsymbol{V}^i}+\sum^{m_2}_{i=m_1+1}{\overline{\boldsymbol{U}^i}},\sum^{m_1}_{i=1}{\boldsymbol{U}^i}+\sum^{m_2}_{i=m_1+1}{\overline{\boldsymbol{V}^i}}\}$. \\
\textbf{Description of Q} \\
From equation \ref{eq:43}, we have Q = $\textrm{max}\{\sum^{m_1}_{i=1}{\boldsymbol{V}^i}+\sum^{m_2}_{i=m_1+1}{\overline{\boldsymbol{U}^i}},\sum^{m_1}_{i=1}{\boldsymbol{U}^i}+\sum^{m_2}_{i=m_1+1}{\overline{\boldsymbol{V}^i}}\}$. Now, $\boldsymbol{U^i}, \boldsymbol{V^i} , \boldsymbol{\overline{U^i}},\boldsymbol{\overline{V^i}}$ for all i are $n \cross 1$ vectors with each element being and element from the matrix $\boldsymbol{A}$ multiplied by some scalar. Since, all elements $A^{ij}$ of the matrix $\boldsymbol{A}$ are drawn from a Gaussian $(0,\frac{1}{m})$ distribution, each row of one of the four vectors $\boldsymbol{V^i},\boldsymbol{U^i}, \overline{\boldsymbol{U^i}}, \overline{\boldsymbol{U^i}} $ is a scalar multiplied to $A^{ij}$. Let $C_1$ be the upper bound of all the scalars multiplied to all the vectors $\boldsymbol{U^i}, \boldsymbol{V^i} , \boldsymbol{\overline{U^i}},\boldsymbol{\overline{V^i}}$ for all i. Now, $C_1A^{ij} \sim N(0, \frac{C_1^2}{m})$. For each term, $\sum^{m_1}_{i=1}{\boldsymbol{V}^i}+\sum^{m_2}_{i=m_1+1}{\overline{\boldsymbol{U}^i}}$ and $\sum^{m_1}_{i=1}{\boldsymbol{U}^i}+\sum^{m_2}_{i=m_1+1}{\overline{\boldsymbol{V}^i}}$ , there are upper bounded by $m_1+m_2$ terms of the form $C_1A^{ij}$. So, the aforementioned terms will be bounded above by a term which has the distribution $N(0,\frac{m_1C_1^2}{m})$. To find  Q, we need to find the maximum between $\sum^{m_1}_{i=1}{\boldsymbol{V}^i}+\sum^{m_2}_{i=m_1+1}{\overline{\boldsymbol{U}^i}}$ and $\sum^{m_1}_{i=1}{\boldsymbol{U}^i}+\sum^{m_2}_{i=m_1+1}{\overline{\boldsymbol{V}^i}}$. To do this, we use the union bound used in example 11.1 from \cite{THW2015}. Since the vectors are of the dimension $n \cross 1$, putting the union bound on the two aforementioned term brings in the quantity $log(n)$ in the structure of Q. From that, we get Q of the form $C_1\sqrt{\frac{(m_1+m_2)log(n)}{m}}$.
\\
\subsubsection{$L_{\infty}$ norm bound on Term 3}
We have , \textbf{Term 3}= $-\dfrac{1}{\sigma}{\sum^{m_3}_{i=m_2+1}{\boldsymbol{A^i}(\dfrac{y_i-\boldsymbol{A^i}\boldsymbol{x^*}}{\sigma})}}$ \\
We know, $y_i \sim N(\boldsymbol{A^i x},\sigma^2) \quad \forall i=1,2,....,m$. \\
Standardising, $z_i=\frac{y_i-\boldsymbol{A^ix}}{\sigma} \quad \forall i=1,2,..,m$ . Hence, $z_i \sim N(0,1) \quad \forall i=1,2,...,m$ 
Diving the matrix $\boldsymbol{A}$ as,
$\boldsymbol{A}^{m \cross n}$=
$
\begin{bmatrix}
\boldsymbol{A_1}^{m_1 \cross n} \\
\boldsymbol{A_2}^{m_2 \cross n} \\
\boldsymbol{A_3}^{m_3 \cross n} \\
\end{bmatrix}
$
Now, $A^{ij}$ : <i,j>th element of $\boldsymbol{A}$ $\sim N(0,\frac{1}{m})$.  Let $\boldsymbol{z_3}:$ standarised measurements w.r.t. $\boldsymbol{A_3}$. \\
Clearly, \textbf{Term 3}= $-\frac{1}{\sigma} \boldsymbol{A^T_3}\boldsymbol{z}_3$. Note that $A^{jT}_3\textbf{z}_3$ is the $j^{th}$ element of the vector $\boldsymbol{A^T_3}\boldsymbol{z_3} \quad \forall j=1,2,..,n$. By the property of linear combination of normal variables, \\
\begin{equation} \label{eq:44}
    A^{jT}_3\textbf{z}_3 \sim N(0, \|\boldsymbol{A_3^j}\|_2^2) \quad \forall j=1,2,...,n
\end{equation}
where, $\|\boldsymbol{A_3^j}\|_2^2=\sum^{m_3}_{i=m_2+1}{(A_3^{ij})^2}$. Now,
\begin{eqnarray} \label{eq:45}
    A_3^{ij} \sim N(0,\frac{1}{m}) \implies \sqrt{m}A_3^{ij} \sim N(0,1) \implies m(A_3^{ij})^2 \sim \chi^2_1 \nonumber \\ \implies m \sum^{m_3}_{i=m_2+1}{(A_3^{ij})^2} \sim \chi_{m_3}^2 \implies \mathcal{E} \,[m\sum^{m_3}_{i=m_2+1}{(A_3^{ij})^2}] = m_3 \nonumber \\ \implies \mathcal{E} \,[\sum^{m_3}_{i=m_2+1}{(A_3^{ij})^2}] = \dfrac{m_3}{m} \quad \forall j=1,2,...,n
\end{eqnarray}
We take the approximation $\|\boldsymbol{A_3^j}\|_2^2 = \frac{m_3}{m} \quad \forall j=1,2,...,n$. Hence,
\begin{eqnarray} \label{eq:46}
    A_3^{jt}\boldsymbol{z}_3 \sim N(0,\frac{m_3}{m}) \implies -A_3^{jt}\boldsymbol{z}_3 \sim N(0,\frac{m_3}{m}) \nonumber \\ \implies \dfrac{-A_3^{jt}\boldsymbol{z}_3}{\sigma} \sim N(0,\frac{m_3}{m\sigma^2})  
\end{eqnarray}
Thus, the Gaussian tail bound is given by,
\begin{equation} \label{eq:47}
    \mathcal{P}\,\left[|\dfrac{\boldsymbol{A_3}^{t}\boldsymbol{z}_3}{\sigma}| \geq u\right] \leq 2 \exp{-\dfrac{u^2\sigma^2m}{2m_3}} 
\end{equation}
The union bound on equation \ref{eq:47} gives us,
\begin{equation} \label{eq:48}
    \mathcal{P}\,\left[\|\dfrac{\boldsymbol{A_3}^{t}\boldsymbol{z}_3}{\sigma}\|_{\infty} \geq u\right] \leq 2 \exp{-\dfrac{u^2\sigma^2m}{2m_3}+\log(n)}
\end{equation}
Equality takes place when , $u=\frac{1}{\sigma}\sqrt{\frac{m_3 \varrho\log(n)}{m}}$, where, $\varrho > 2$. Thus, 
\begin{eqnarray} \label{eq:49}
    \mathcal{P}\,\left[\|\dfrac{\boldsymbol{A_3}^{t}\boldsymbol{z}_3}{\sigma}\|_{\infty} \geq \frac{1}{\sigma}\sqrt{\frac{m_3\varrho\log(n)}{m}}\right] \leq 2\exp{-\frac{1}{2}(\varrho-2)\log(n)} \nonumber \\ \implies \|\textbf{Term 3}\|_{\infty} \geq  \frac{1}{\sigma}\sqrt{\frac{m_3\varrho\log(n)}{m}} \\ \textrm{with prob.}\, 2\exp{-\frac{1}{2}(\varrho-2)\log(n)} \label{eq:50}
\end{eqnarray}
\subsection{$L_{\infty}$ norm Bound for $\nabla L$}
Let $\vartheta_1=\textrm{Term 3}$ and $\vartheta_2=\textrm{Term 2 - Term 1}$.\\
From the \textbf{Reverse Triangle Inequality} on equations \ref{eq:43} and \ref{eq:49} , we have,
\begin{eqnarray} \label{eq:51}
    \|\vartheta_1+\vartheta_2\|_{\infty} \geq |\,\|\vartheta_1\|_{\infty} - \|\vartheta_2\|_{\infty} \,| \nonumber \\ \implies 
    \|\textbf{-Term 1+Term 2+Term 3}\|_{\infty} \geq |\frac{1}{\sigma}\sqrt{\frac{m_3\varrho\log(n)}{m}}-\frac{Q}{\sigma}| \nonumber \\ \implies \|\boldsymbol{\nabla L}\|_{\infty} \geq \frac{1}{\sigma}|\sqrt{\frac{m_3\varrho\log(n)}{m}}-Q| 
\end{eqnarray}
with probability $2\exp{-\frac{1}{2}(\varrho-2)\log(n)}$.\\
Again, from the \textbf{Triangle Inequality} on  \ref{eq:43} and \ref{eq:49}, we have,
\begin{eqnarray} \label{eq:52}
    \|\vartheta_1+\vartheta_2\|_{\infty} \leq \|\vartheta_1\|_{\infty} + \|\vartheta_2\|_{\infty} \nonumber \\ \implies
    \|\textbf{-Term 1+Term 2+Term 3}\|_{\infty} \leq \frac{1}{\sigma}\sqrt{\frac{m_3\varrho\log(n)}{m}}+\frac{Q}{\sigma} \nonumber \\
    \implies \|\boldsymbol{\nabla L}\|_{\infty} \leq \frac{1}{\sigma}\{\sqrt{\frac{m_3\varrho\log(n)}{m}}+Q\}
\end{eqnarray}
\\
with probability $2\exp{-\frac{1}{2}(\varrho-2)\log(n)}$. 
where, $Q \triangleq \textrm{max}\{\sum^{m_1}_{i=1}{\boldsymbol{V}^i}+\sum^{m_2}_{i=m_1+1}{\overline{\boldsymbol{U}^i}},\sum^{m_1}_{i=1}{\boldsymbol{U}^i}+\sum^{m_2}_{i=m_1+1}{\overline{\boldsymbol{V}^i}}\}$. \\
This upper bound will be useful in the final performance bounds. 

\subsection{Theorem 4: Upper bound on the Reconstruction Error}
From Theorem-1 of \cite{Negahban2012}, given a $\lambda \geq 2R^{*}(\nabla L(\boldsymbol{y},\boldsymbol{Ax};\tau))$ , for any optimal solution $\widehat{\boldsymbol{x}}_{\lambda}$ with regulariser $\lambda$ , the reconstruction error of the cost function satisfies the upper bound ($\boldsymbol{x}^*$) :-
\begin{equation} \label{eq:53}
    \|\widehat{\boldsymbol{x}}_{\lambda}-\boldsymbol{x}^{*}\|_2^2 \leq 9 \frac{\lambda^2}{\kappa_L^2} \psi^2(\overline{M}) \,+\, \frac{\lambda}{\kappa_L}[2\tau_{L}^2(\boldsymbol{x}^{*})\,+\,4R(\boldsymbol{x}^*_{\overline{M}})]
\end{equation}
where, $R(.)$ is the regularisation function , $R^*(.)$ is the dual of the regularisation function, $\psi^2(\overline{M}) = \sup_{v \in \mathbb{R}}{\frac{R(\boldsymbol{v})}{\|\boldsymbol{v}\|_2}}$ and $\boldsymbol{x}^*_{\overline{M}}$ is all the elements except the s largest elements of vector \textbf{x} as defined in \cite{Negahban2012}. \\
In our model, $R(\boldsymbol{x})=\|\boldsymbol{x}\|_1$ and $R^{*}(\boldsymbol{x})=\|\boldsymbol{x}\|_{\infty}$ . Since the true signal $\boldsymbol{x}$ is assumed to be strictly sparse , $R(\boldsymbol{x}^*_{\overline{M}})=0$. Also,  $\psi^2(\overline{M}) = s$, where $s$ is the sparsity of the original signal $\boldsymbol{x}$ \cite{THW2015}. Now, from the upper bound for $\|\nabla L\|_{\infty}$ in \ref{eq:51}, we take $\lambda = 2({\frac{1}{\sigma}}\sqrt{\frac{m_3\log{n}\varrho}{m}}+\frac{Q}{\sigma})$. where $\varrho > 2$ and Q is as shown in Eqn. \ref{eq:43}. Hence, our upper bound is given by,
\begin{eqnarray} \label{eq:54}
    \|\widehat{\boldsymbol{x}}_{\lambda}-\boldsymbol{x}^{*}\|_2^2 \leq 9 \{\frac{2}{\sigma}(\sqrt{\frac{m_3\log({n})\varrho}{m}}+Q)\}^2 \cross \{\frac{2\sigma^2}{\gamma}\}^2 \cross s \nonumber \\
    = 144s\{\sqrt{\frac{m_3\log({n})\varrho}{m}}+Q\}^2\frac{\sigma^2}{\gamma^2}
\end{eqnarray} \\
where $Q \triangleq \textrm{max}\{\sum^{m_1}_{i=1}{\boldsymbol{V}^i}+\sum^{m_2}_{i=m_1+1}{\overline{\boldsymbol{U}^i}},\sum^{m_1}_{i=1}{\boldsymbol{U}^i}+\sum^{m_2}_{i=m_1+1}{\overline{\boldsymbol{V}^i}}\}$. This proves Theorem 4. As described in Section 5.6, $Q$ is of the order $O(\sqrt{\frac{(m_1+m_2)log(n)}{m}})$. Note that the range of values in the signal $\boldsymbol{x}$ is from $\alpha$ to $\beta$, both of which could potentially have large absolute value. The terms $\boldsymbol{V^i},\boldsymbol{U^i}, \overline{\boldsymbol{U^i}}, \overline{\boldsymbol{U^i}} $ are either of the form $\frac{\phi()}{\Phi()}$ or $\frac{\phi()}{1-\Phi()}$. Hence, these terms can be really large. So, the coefficient $C_1$ in Q would also be very large. Consequently, the terms $m_1+m_2$ dominates in the upper bound presented in Eqn. \ref{eq:54}, i.e. with increase in the saturated measurements the upper bound in the reconstruction error becomes looser.
\bibliographystyle{plain}
\bibliography{refs}
\end{document}